%% file: main.tex
\documentclass[10pt,twocolumn,letterpaper]{article}

\usepackage{cvpr}              


\definecolor{cvprblue}{rgb}{0.21,0.49,0.74}
\usepackage{multirow}
\usepackage{booktabs}
\usepackage{colortbl}       
\usepackage{amsfonts}       
\usepackage{amssymb} 
\usepackage{graphicx}
\usepackage{wrapfig}
\usepackage{amsmath}
\usepackage[accsupp]{axessibility} 
\usepackage[dvipsnames,table]{xcolor}
\usepackage{threeparttable}
\usepackage{lscape}
\usepackage{url} 
\usepackage[figuresright]{rotating}
\definecolor{sky}{RGB}{44, 173, 255}
\usepackage[pagebackref,breaklinks,colorlinks,allcolors=cvprblue]{hyperref}
\hypersetup{
  colorlinks=true,
  linkcolor=cvprblue,  
  citecolor=cvprblue,   
  urlcolor=cvprblue     
}

\usepackage{multirow,makecell}

\usepackage[utf8]{inputenc} 
\usepackage[T1]{fontenc}    
\usepackage{url}            
\usepackage{booktabs}       
\usepackage{amsfonts}       
\usepackage{nicefrac}       
\usepackage{microtype}      
\usepackage{xcolor}    
\usepackage{listings}

\def\paperID{} 
\def\confName{CVPR}
\def\confYear{2026}

\title{\textcolor{sky}{Skyra}: AI-Generated Video Detection via Grounded Artifact Reasoning}

\author{Yifei Li, Wenzhao Zhen$^{\dag}$, Yanran Zhang, Runze Sun, Yu Zheng, Lei Chen, Jie Zhou, Jiwen Lu$^{*}$\\
Department of Automation, Tsinghua University\\
Project Page:  \url{https://joeleelyf.github.io/Skyra}
}

\begin{document}

\setlength{\abovedisplayskip}{1pt}
\setlength{\belowdisplayskip}{1pt}
\setlength{\abovedisplayshortskip}{1pt}
\setlength{\belowdisplayshortskip}{1pt}

\maketitle
\begingroup
    \renewcommand{\thefootnote}{}
    \footnotetext{\noindent\textdagger{} Project leader.}
    \footnotetext{\noindent* Corresponding author.}
    \addtocounter{footnote}{-2}
\endgroup

\input{sec/0_Abstract}    
\input{sec/1_Intro}
\input{sec/2_Related}

\input{sec/3_Dataset}
\input{sec/4_Method}

\input{sec/5_Experiments}
\input{sec/6_Conclusion}
\input{sec/7_Acknowledge}

\clearpage

{
    \small
    \bibliographystyle{ieeenat_fullname}
    \bibliography{main}
}

\input{sec/X_suppl}

\end{document}

%% file: sec/0_Abstract.tex
\begin{abstract}
The misuse of AI-driven video generation technologies has raised serious social concerns, highlighting the urgent need for reliable AI-generated video detectors. However, most existing methods are limited to binary classification and lack the necessary explanations for human interpretation. 
In this paper, we present \textbf{Skyra}, a specialized multimodal large language model (MLLM) that identifies human-perceivable visual artifacts in AI-generated videos and leverages them as grounded evidence for both detection and explanation. 
To support this objective, we construct \textbf{ViF-CoT-4K} for Supervised Fine-Tuning (SFT), which represents the first large-scale AI-generated video artifact dataset with fine-grained human annotations.
We then develop a two-stage training strategy that systematically enhances our model's spatio-temporal artifact perception, explanation capability, and detection accuracy. 
To comprehensively evaluate Skyra, we introduce \textbf{ViF-Bench}, a benchmark comprising 3K high-quality samples generated by over ten state-of-the-art video generators. Extensive experiments demonstrate that Skyra surpasses existing methods across multiple benchmarks, while our evaluation yields valuable insights for advancing explainable AI-generated video detection. 
Our code, models, and datasets are publicly available at \href{https://joeleelyf.github.io/Skyra}{https://joeleelyf.github.io/Skyra}.
\end{abstract}

%% file: sec/1_Intro.tex
\section{Introduction}
\label{sec:intro}
With the rapid evolution of diffusion-based~\cite{blattmann2023stable, peebles2023scalable, yang2024cogvideox, wan2025wan, kong2024hunyuanvideo} and multimodal generative models~\cite{klingai, google_veo3, Sora2_2025}, synthetic videos now achieve unprecedented levels of authenticity, enabling anyone to produce photorealistic content from simple text prompts or reference images. While this progress reshapes entertainment, communication, and design, its misuse poses growing threats to social safety~\cite{xu2023combating, shoaib2023deepfakes}. 

\begin{figure}[t]
    \centering
    \includegraphics[width=0.475\textwidth]{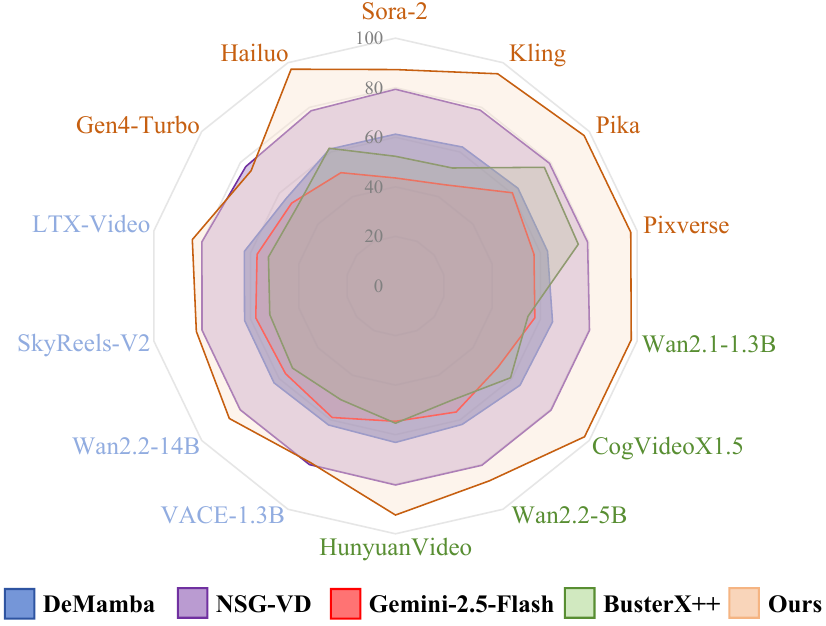}
    \vspace{-7mm}
    \caption{
        Performance on ViF-Bench. Our method outperforms both binary and existing MLLM-based detectors.
    }
    \vspace{-5mm}
    \label{fig:performance}
\end{figure}

\begin{figure*}[t]
    \centering
    \includegraphics[width=1\textwidth]{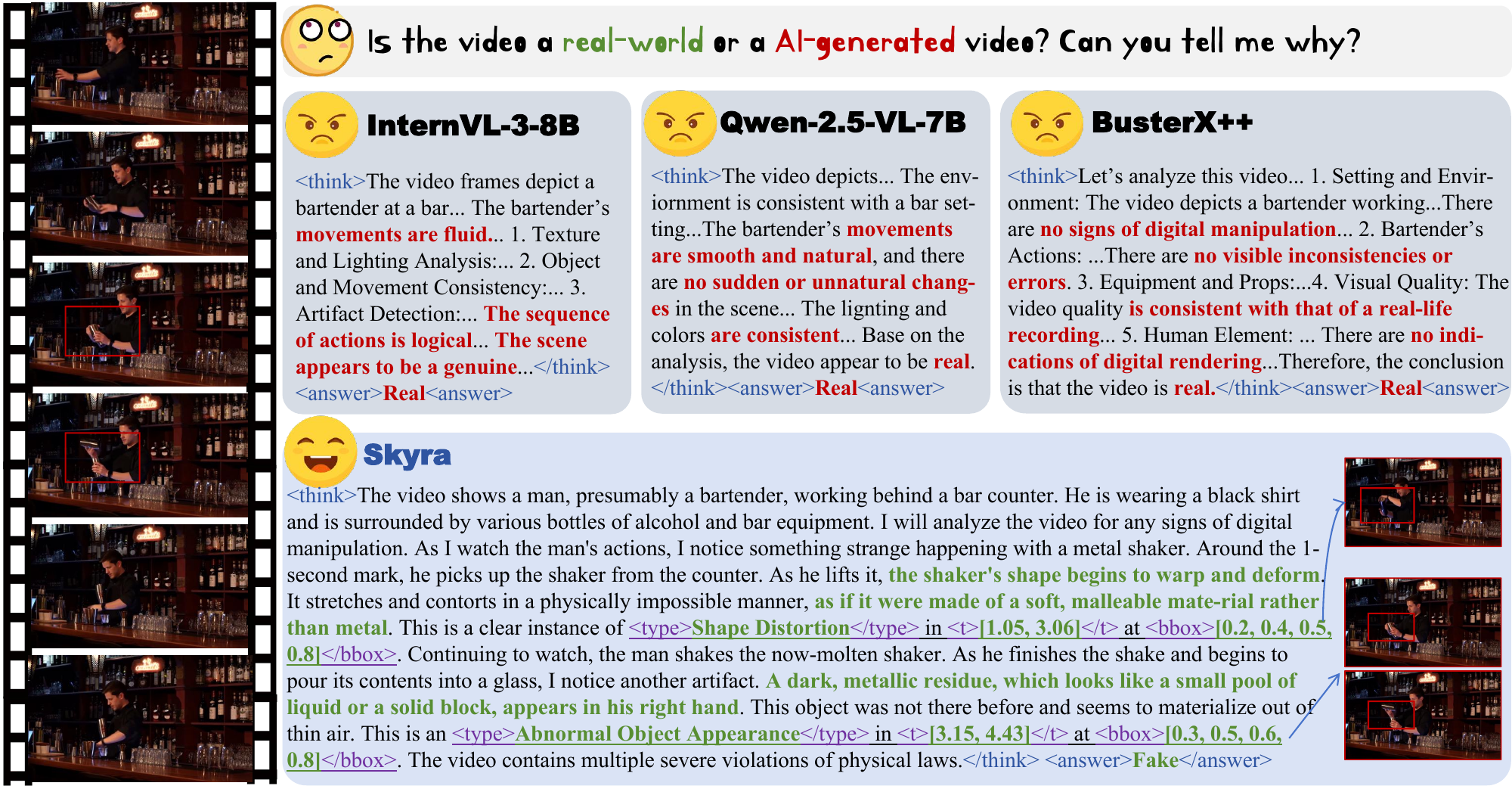}
    \vspace{-7mm}
    \caption{
        \textbf{Skyra} leverages human-perceivable artifacts in AI-generated videos as grounded evidence for detection and explanation. Compared to off-the-shelf MLLMs and previous MLLM-based detectors, Skyra demonstrates superior artifact perception and detection capabilities.
    }
    \vspace{-5mm}
    \label{fig:teaser}
\end{figure*}

Driven by this urgent need, the community has developed several detection models~\cite{bai2024ai,chen2024demamba,zheng2025d3,zhang2025physics,interno2025ai,park2025vidguard}, datasets~\cite{chen2024demamba,chen2025genworld}, and benchmarks~\cite{chen2024demamba,ni2025genvidbench,chen2025genworld} to detect AI-generated videos. 
The rise of multimodal large language models (MLLMs)~\cite{llava, li2023blip, hurst2024gpt, QWen2.5VL,qwen3vl2025} has attracted the attention of AI-generated content detection researchers due to their capacity for interpretable reasoning~\cite{zhang2024common,li2025fakebench,guo2025rethinking,chen2024x2,tan2025veritas}. 
Still, we empirically find that even state-of-the-art (SoTA) general MLLM~\cite{QWen2.5VL,google_gemini2.5,openai_gpt4.1} achieves near-random performance in identifying AI-generated videos, and fails to capture human-perceivable artifacts, even with carefully designed chain-of-thought (CoT)~\cite{wei2022chain} prompts. While recent works such as BusterX++~\cite{wen2025busterx,wen2025busterx++} attempt to adapt pretrained MLLMs for AI-generated video detection, the resulting model acts more as general content descriptors and overemphasizes superficial cues (e.g., visual quality, lighting) while neglecting the intrinsic, physics-violating artifacts that humans rely on to identify AI-generated videos (Figure~\ref{fig:teaser}).  DAVID-XR1~\cite{gao2025david} advancing the field by introducing human annotations of AI-generated video artifacts. However, the classification taxonomy of their annotations is vague, with the number of valid samples being limited, and the resulting model's performance far from satisfactory. 

To overcome these limitations, we introduce \textbf{Skyra}, a specialized multimodal large language model AI-generated video detection via grounded artifact reasoning, which identifies artifacts and leverages them as spatio-temporally grounded evidence. 
As shown in Figures~\ref{fig:performance} and \ref{fig:teaser}, Skyra achieves substantially higher detection accuracy while providing fine-grained, human-interpretable artifact localization, consistently outperforming both binary classifiers and prior MLLM-based approaches. 
Recognizing that off-the-shelf MLLMs lack sensitivity to subtle generative artifacts, we construct the first large-scale human-annotated AI-generated video artifacts dataset, \textbf{ViF-CoT-4K}, which enables supervised fine-tuning and yields \textbf{Skyra-SFT}. We further propose a second-stage reinforcement learning procedure that pushes forward the model’s ability to mine discriminative artifacts, producing additional gains in both detection and explanation quality, resulting in our final model \textbf{Skyra-RL}. To comprehensively evaluate the ability of existing methods, we release \textbf{ViF-Bench}, which includes high-quality samples generated by over ten latest models, with real and fake samples aligned in both semantics and formats, mitigating shortcut signals, and providing a fair testbed of artifact-based detection.

%% file: sec/2_Related.tex
\begin{figure*}[ht]
    \centering
    \includegraphics[width=1\textwidth]{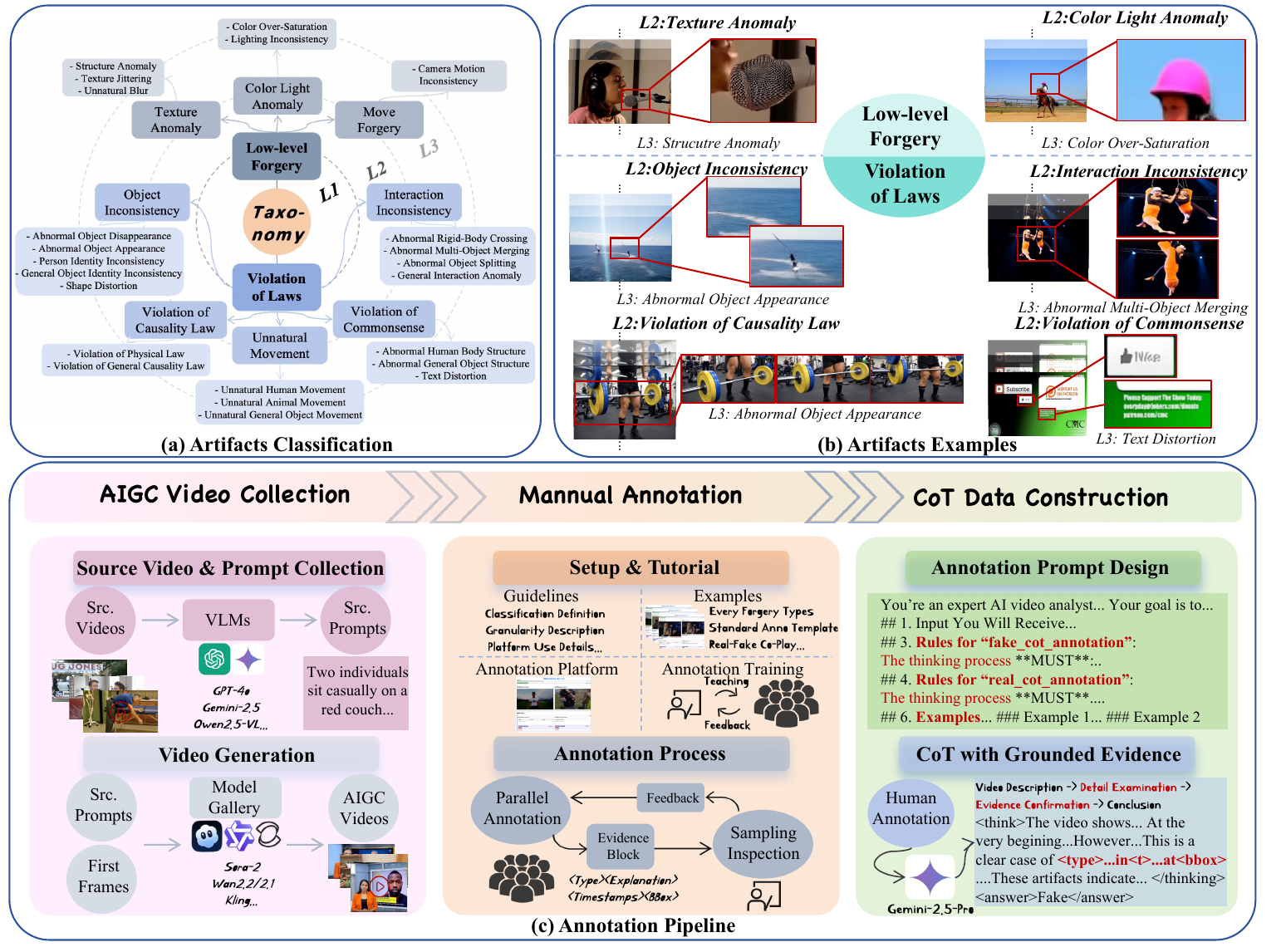}
    \vspace{-7mm}
    \caption{
        Overview of the ViF-CoT-4K dataset. \textbf{(a)} The hierarchical taxonomy of AI-generated video artifacts. \textbf{(b)} Visual examples of artifacts under our taxonomy. \textbf{(c)} Construction pipeline of ViF-CoT-4K dataset, including authentic data collection and AI-generated video collection, manual annotation, and the step-by-step chain-of-thought explanation data construction process. 
    }
    \vspace{-5mm}
    \label{fig:dataset_overview}
\end{figure*}

\section{Related Work}
\label{sec:related}
\noindent\textbf{AI-Generated Video Detection.} AI-generated video detection has largely focused on binary classification~\cite{bai2024ai, ma2024decof, chen2024demamba, interno2025ai, chen2025genworld}, relying on visual artifact detection in synthetic content. 
Early methods like AIGVDet~\cite{bai2024ai} and DeCoF~\cite{ma2024decof} exploit spatio-temporal features and frame consistency to detect discrepancies, achieving strong performance on curated datasets such as GVD~\cite{bai2024ai} and GVF~\cite{ma2024decof}. Recent works, including DeMamba~\cite{chen2024demamba}, D3~\cite{zheng2025d3}, ReStraV~\cite{interno2025ai}, and NSG-VD~\cite{zhang2025physics}, explore more discriminative and robust feature spaces, demonstrating success on updated benchmarks like GenVideo~\cite{chen2024demamba} and GenVidBench~\cite{ni2025genvidbench}. However, these approaches lack interpretability, leaving the detection process opaque and limiting their applicability in scenarios that require manual verification.

The emergence of MLLMs~\cite{llava, Video-llama, chen2024internvl, zhang2024internlm, Qwen2VL, QWen2.5VL, team2023gemini, hurst2024gpt} has enabled more explainable detection~\cite{zhang2024common, guo2025rethinking, xu2024fakeshield, sun2024forgerysleuth, park2025vidguard}, providing both predictions and reasoning processes. X2-DFD~\cite{chen2024x2} and VERITAS~\cite{tan2025veritas} demonstrate MLLM effectiveness in deepfake face detection through pattern-aware reasoning and feature enhancement. MLLMs have also succeeded in AIGC image detection, with frameworks like FakeVLM~\cite{wen2025spot} and LEGION~\cite{kang2025legion} enabling fine-grained artifact identification, and UniGenDet~\cite{zhang2026unigendet} unifying image generation and generated-image detection within a single co-evolutionary framework. For video content, MLLM-based methods remain nascent. IVY-Fake~\cite{zhang2025ivy}, DAVID-XR1~\cite{gao2025david}, and BusterX++~\cite{wen2025busterx,wen2025busterx++} pioneer explainable video detection, providing interpretable reasoning on motion, texture, and temporal artifacts. Concurrently, DeepTraceReward~\cite{fu2025learning} studies human-perceived fakeness in AI-generated videos through multimodal LLMs, proposing evaluation metrics for explanation quality including spatial and temporal localization accuracy. Despite progress, current methods struggle with complex temporal dynamics and fine-grained reasoning, often relying on MLLM-generated annotations~\cite{zhang2025ivy, team2023gemini} or basic fine-tuning~\cite{zhang2025ivy, gao2025david}. In contrast, our approach leverages high-quality human annotations and precise spatio-temporal supervision to enhance artifact perception and reasoning.

\noindent\textbf{Multimodal Large Language Models for Video.} Recent advances in video MLLMs have yielded specialized architectures for video processing and reasoning. Video-ChatGPT~\cite{maaz2023video} integrates a video-adapted visual encoder with an LLM for detailed understanding through conversation. Video-LLaMA~\cite{Video-llama} employs multimodal encoders for spatio-temporal reasoning, integrating audio and video for enhanced comprehension.
Meanwhile, general vision-language models have also demonstrated strong video capabilities. Qwen2-VL~\cite{Qwen2VL} introduces dynamic-resolution tokenization and unified image-video encoding. Recent models like Seed1.5VL~\cite{guo2025seed1}, InternVL3.5~\cite{wang2025internvl3}, and Qwen3VL~\cite{qwen3vl2025} further advance visual-temporal feature integration, achieving strong performance on diverse benchmarks~\cite{li2024mvbench, wang2025lvbench, wu2024longvideobench, niu2025ovo}.

Despite these advances, base MLLMs exhibit limited reasoning without task-specific fine-tuning. Post-training strategies, particularly reinforcement learning (RL)~\cite{su2025thinking, tang2025video}, have emerged to address this gap. 
In the image domain, OpenThinkIMG~\cite{su2025openthinkimg} and DeepEyes~\cite{zheng2025deepeyes} incorporate RL to enhance visual reasoning through structured, multi-stage processes. For video understanding, Video-R1~\cite{feng2025video} introduces T-GRPO for temporally consistent reasoning, while LongVILA-R1~\cite{chen2025scaling} integrates RL with large-scale reasoning tasks to support longer inputs. Recent methods~\cite{fu2025love, yan2025videochat, he2025framethinker} enhance reasoning by incorporating tool use into trajectories, achieving state-of-the-art performance. These RL-based strategies are essential for enhancing MLLMs' perception~\cite{liu2025seg,zhang2025chain,lai2025mini} and reasoning~\cite{hu2024visual,zhang2025thyme,guo2025geovlmath}, particularly for challenging video understanding tasks~\cite{tang2025video}.

%% file: sec/3_Dataset.tex
\begin{figure*}[ht]
    \centering
    \includegraphics[width=1\textwidth]{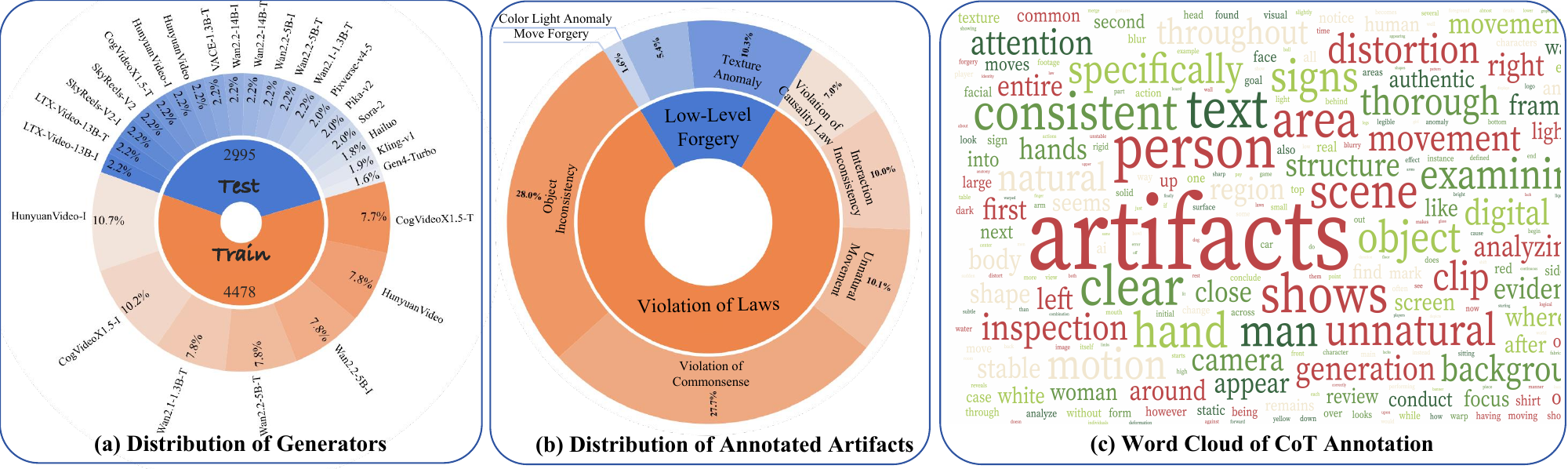}
    \vspace{-7mm}
    \caption{
        Statistics of the ViF-CoT-4K and ViF-Bench. \textbf{(a)} Distribution of samples generated by different generators in ViF-CoT-4K (train) and ViF-Benchmark (test) set. \textbf{(b)} Distribution of artifacts types in ViF-CoT-4K. Detailed proportion is provided in the Appendix. \textbf{(c)} Word cloud of the CoT annotations in ViF-CoT-4K.
    }
    \vspace{-5mm}
    \label{fig:dataset_statistics}
\end{figure*}

\section{ViF Dataset}
\label{sec:dataset}
With rapid advances in AI video generation technologies~\cite{yang2024cogvideox,wan2025wan,Sora2_2025,google_veo3}, numerous datasets~\cite{chen2024demamba,wen2025busterx,chen2025genworld} and benchmarks~\cite{chen2024demamba, ni2025genvidbench,chen2025genworld} have emerged for detection research. 
However, existing datasets face three key limitations: \textbf{(1) Significant Real-Fake Discrepancy}: In datasets like \cite{chen2024demamba, ni2025genvidbench,chen2025genworld}, real videos exhibit 2-3× higher duration and FPS than fake counterparts. Moreover, Domain and style distributions also differ substantially~\cite{chen2025genworld}, enabling models to exploit spurious correlations through shortcut learning~\cite{park2025vidguard}. \textbf{(2) Limited Diversity and Authenticity}: Most datasets include few~\cite{park2025vidguard} or outdated generative models~\cite{chen2024demamba}. 
For example, VidGuard-R1~\cite{park2025vidguard} relies solely on HunyuanVideo~\cite{kong2024hunyuanvideo} and CogVideoX~\cite{yang2024cogvideox}, while GenVidBench~\cite{ni2025genvidbench} mainly includes models released over two years ago~\cite{hong2022cogvideo,wang2023modelscope,blattmann2023stable,khachatryan2023text2video}. This homogeneity limits utility for real-world applications, where models such as Sora-2~\cite{Sora2_2025}, Wan2.2~\cite{wan2025wan}, and Kling~\cite{klingai} now generate highly realistic visual content. 
\textbf{(3) Lack of Detailed Artifact Annotations}: Mainstream detection methods and datasets\cite{chen2024demamba,ni2025genvidbench,zhang2025physics} focus solely on binary classification without detection rationale. While IVY-Fake~\cite{zhang2025ivy} and VidGuard-R1~\cite{park2025vidguard} attempt explanations by prompting general MLLMs~\cite{google_gemini2.5}, they lack a systematic artifact taxonomy and grounded localization.

To address these limitations, we propose a dataset and benchmark construction pipeline with fine-grained manual annotations, as illustrated in Figure~\ref{fig:dataset_overview}. This section presents the statistical analysis (§\ref{subsec:statistics}), our artifact taxonomy (§\ref{subsec:taxonomy}), and dataset construction process(§\ref{subsec:construction}).

\subsection{Dataset Statistics}
\label{subsec:statistics}
We provide a statistical report of our constructed dataset and benchmark in Figure~\ref{fig:dataset_statistics}. Distribution of video generators and numbers of different annotated artifact types, and word cloud of the CoT explanation are exhibited. Additional details, including video statistics, annotation guidelines, and complete CoT prompts, are provided in the Appendix. 

\subsection{Artifact Taxonomy}
\label{subsec:taxonomy}
A comprehensive and unambiguous artifact taxonomy is essential for high-quality manual annotation and model reasoning. Prior MLLM-based detection methods either lack~\cite{wen2025busterx, wen2025busterx++}  or provide vague, coarse-grained taxonomies~\cite{gao2025david, zhang2025ivy}. 
Categories such as ``Space Anomaly'' and ``Spatial Relationships''~\cite{gao2025david, zhang2025ivy} lack granularity and cause ambiguity.

To address these limitations, we propose a hierarchical taxonomy for fine-grained classification of human-perceivable artifacts. 
Our taxonomy comprises three layers. 
\textbf{Layer 1 (L1)} defines two high-level categories: \textit{Low-level forgery} (perceptual quality artifacts) and \textit{Violation of Laws} (physical and logical inconsistencies). 
\textbf{Layer 2 (L2)} refines these into eight categories: \textit{Low-level Forgery} includes \textit{Color/Light Anomaly}, \textit{Texture Anomaly}, and \textit{Motion Forgery}; \textit{Violation of Laws} includes \textit{Object Inconsistency}, \textit{Interaction Inconsistency}, \textit{Violation of Causality}, \textit{Violation of Commonsense}, and \textit{Unnatural Movement}. 
\textbf{Layer 3 (L3)} provides the most fine-grained, observable artifacts. 
For instance, the \textit{Object Inconsistency} in L2 divides into \textit{Abnormal Object Disappearance}, \textit{Abnormal Object Appearance}, and \textit{Person Identity Inconsistency}. This hierarchical structure progresses from abstract concepts to specific observable patterns. The complete taxonomy with examples is shown in Figure~~\ref{fig:dataset_overview}. 

\begin{figure*}[t]
    \centering
    \includegraphics[width=1\textwidth]{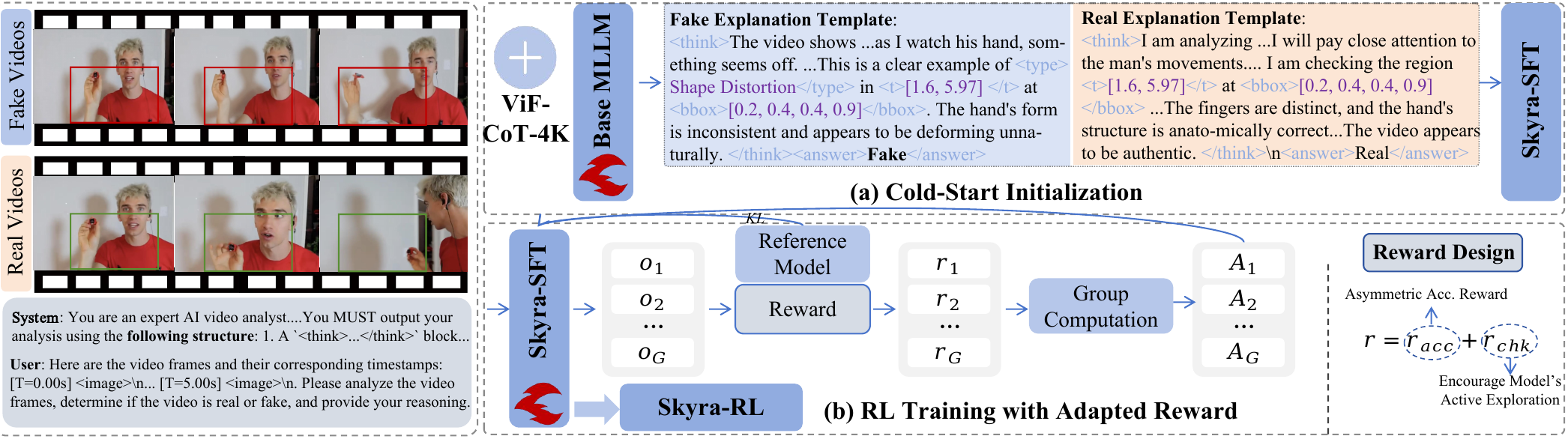}
    \vspace{-7mm}
    \caption{
        Overview of Skyra. We leverage a two-stage training pipeline to improve Skyra's artifacts perception and detection capabilities: \textbf{(a)} cold-start initialization with balanced fake and real explanation response templates to endow the base model with basic AI-generated artifacts perception capability. \textbf{(b)} reinforcement learning with adapted rewards to encourage the model's self-driven visual probe process.
    }
    \vspace{-5mm}
    \label{fig:method_overview}
\end{figure*}

\subsection{Dataset and Benchmark Construction}
\label{subsec:construction}
\noindent\textbf{Real and AI-Generated Videos Collection.} We sample around 3.5K real videos from Panda-70M~\cite{chen2024panda} and 1.5K from Kinetics-400~\cite{kay2017kinetics}, covering diverse content types including real-life recordings, TV shows, and human actions. 
We also include high-resolution videos from HD-VILA-100M~\cite{xue2022advancing} in our benchmark to test the generalization ability. 
We then utilize a variety of MLLMs~\cite{hurst2024gpt,openai_gpt4.1,google_gemini2.5,QWen2.5VL,chen2024internvl} to generate detailed video descriptions, which are transformed into prompts for generation models after manual quality inspection. 
These prompts drive text-to-video (T2V) models~\cite{wan2025wan,yang2024cogvideox,kong2024hunyuanvideo,jiang2025vace,chen2025skyreels,hacohen2024ltx,runwayml2025gen4,hailuo2025hailuo02,pika2025pikaart,pixverse2025pixverse}. 
For image-to-video (I2V) generation~\cite{kong2024hunyuanvideo,chen2025skyreels,hacohen2024ltx}, we extract the first frame from real videos as conditions. An automatic filtering pipeline using GPT-4o-mini~\cite{openai2024gpt4omini} ensures semantic consistency between AI-generated and real videos, addressing Limitation (1). 

To address Limitation (2), we incorporate diverse state-of-the-art generators spanning open-source and commercial domains. For training, we use Wan2.2-TI2V-5B~\cite{wan2025wan}, Wan2.1-T/I2V-1.3B~\cite{wan2025wan}, CogVideoX-1.5-5B~\cite{yang2024cogvideox}, and HunyuanVideo~\cite{kong2024hunyuanvideo}. For evaluation, we include recent models like Wan2.2-T/I2V-A14B~\cite{wan2025wan}, LTX-Video-13B~\cite{hacohen2024ltx}, MiniMax-Hailuo~\cite{hailuo2025hailuo02}, and Sora-2~\cite{Sora2_2025}. See Table~\ref{tab:main_results} for details. 

\noindent\textbf{Fine-grained Manual Annotation.} We collaborate with domain experts to develop detailed annotation guidelines and build an easy-to-use annotation platform. Professional annotators familiar with video generation models identify all visible artifacts (\textit{fake evidence}) in AI-generated videos, annotating: (1) artifact \textbf{Type} from our taxonomy, (2) \textbf{Textual Explanation}, (3) temporal-spatial localization via \textbf{Timestamps} and \textbf{Bounding Boxes}. Notably, we display AI-generated videos alongside real counterparts, prompting annotators to identify corresponding \textit{real evidence} in real videos for each \textit{fake evidence}. 
This helps to validate that artifacts are truly generation-induced rather than compression-related degradation. Multiple review cycles ensure annotation quality and inter-annotator consistency. 

\noindent\textbf{Chain-of-Thought (CoT) Annotation.} Chain-of-thought reasoning improves MLLM performance on complex visual tasks~\cite{wei2022chain}. While precise, our manual labels lack step-by-step reasoning valuable for model training. For each AI-generated video, we feed artifact \textbf{Type}, \textbf{Textual Explanation}, \textbf{Timestamps}, and \textbf{Bounding Boxes} aligned with sampled frames to Gemini-2.5-Pro~\cite{google_gemini2.5}. 
For real videos, we substitute \textit{fake evidence} with \textit{real evidence} as model input. 
To improve the quality of annotation generated by Gemini-2.5-Pro, we employ two prompt-engineering strategies.  (1) \textbf{Self-Curation}: we instruct the model to follow an \textit{observe-understand-draft-review-conclude} process, grounding CoT in visual details; (2) \textbf{In-Context Learning}: we provide detailed definitions and carefully crafted CoT examples for each artifact type, filtering mismatched annotations. 

%% file: sec/4_Method.tex
\section{Skyra}
\label{sec:method}
In this section, we analyze the characteristics of the AI-generated video detection and explanation, examine the challenges of applying off-the-shelf MLLMs to this task, and present our design motivation (Section~\ref{sec:analysis}). 
We then introduce our two-stage training strategies as depicted in Figure~\ref{fig:method_overview}, i.e., supervised fine-tuning to endow the model with basic detection and explanation capabilities (Section~\ref{sec:sft}), and reinforcement learning to enhance its ability to perceive and reason about AI-generated artifacts (Section~\ref{sec:rl}).

\subsection{Analysis of the AI-Generated Video Detection and Explanation Task}
\label{sec:analysis}
\noindent\textbf{How Humans Identify AI-Generated Videos.} Conventional AIGC detection approaches typically extract handcrafted~\cite{zheng2025d3,interno2025ai,zhang2025physics} or learned features~\cite{bai2024ai,chen2024demamba,chen2025genworld} from generative samples and perform binary classification in this feature space. However, this paradigm often devolves into a continuous adversarial cycle between detectors and generators: as new generative models emerge, previous discriminative features may lose effectiveness. This results in detectors that lack generalizability and remain fragile when encountering unseen samples~\cite{zou2025survey}. 

To move beyond this limitation, we examine how humans identify AI-generated videos. 
Humans first perceive the overall semantic and temporal context, then actively search for spatio-temporal inconsistencies, such as abrupt object disappearance, unnatural motions, or implausible scene transitions that reveal synthetic content. 
Through continuous interaction with the real world, humans develop a grounded understanding of physical and temporal coherence~\cite{spelke1994initial,shepard1994perceptual,spelke2007core}, enabling intuitive recognition of violations. 
We refer to such cues as \textbf{\textit{intrinsic evidences}}, as they are universal, model-agnostic indicators of deviation from real-world dynamics.

\noindent\textbf{Challenges of Adapting MLLMs for AI-Generated Video Detection.} Pre-trained on large-scale vision-language datasets, MLLMs have acquired a foundational understanding of the real world to some extent. Inspired by prior work~\cite{zhang2024common}, we explored directly prompting off-the-shelf MLLMs for AIGC video detection. 
However, both direct question-answer prompting and carefully designed chain-of-thought (CoT) strategies yield limited accuracy, often below 60\% on our benchmarks (Table.~\ref{tab:main_results}). 

Our experimental analysis reveals two key issues: (1) most existing MLLMs~\cite{wang2025internvl3,QWen2.5VL,Video-llama,openai_gpt4.1}, even with explicit step-by-step CoT guidance, struggle to uncover subtle spatiotemporal forgery cues; and (2) some models~\cite{google_gemini2.5} misinterpret natural video degradations (e.g., compression artifacts, motion blur) as forgery signs, leading to false positives on real videos. Detailed examples are provided in the Appendix.  
These observations motivate our approach to emulate human reasoning, i.e., enhancing the model ability to discover essential forgery cues while incorporating self-verification to re-examine suspected regions in real videos, improving both precision and confidence.

\subsection{Cold-start Initialization}
\label{sec:sft}

\textbf{Response Template Design.} 
We require the model to ground its judgment on careful video inspection and explicitly expose its reasoning process. This chain-of-thought (CoT) supervision is essential for improving accuracy and credibility (see Ablation-II in Section~\ref{sec:ablation}). 
The model follows the format $F_{outer}$:
\vspace{-2mm}
\begin{verbatim}
<thinking>[thinking process]</thinking>
<answer>[Fake / Real]</answer>
\end{verbatim}
\vspace{-2mm}

For \textbf{fake} videos, we guide attention to forged regions using $F_{fake}$:
\texttt{<type>[Forgery Type]</type> in <t>[t\_start, t\_end]</t> at <bbox>[x\_min, y\_min, x\_max, y\_max]</bbox>}. 

For \textbf{real} videos, the model instead inspects suspected regions with the same temporal–spatial tags $F_{real}$: \texttt{<t>[t\_start, t\_end]</t> at <bbox>[x\_min, y\_min, x\_max, y\_max]</bbox>}

\input{tables/main_results}

\noindent\textbf{Training Process.} 
We fine-tune Qwen2.5-VL-7B on our ViF-CoT-4K dataset. 
Text $t$ and video $v$ are encoded by pretrained textual and visual encoders, fused, and fed into the decoder for autoregressive generation. 
Given the ground-truth response sequence $y^{*} = (y^{*}_1, \ldots, y^{*}_T)$, the model is trained with standard cross-entropy loss:
\vspace{-0.5mm}
\begin{equation}
\mathcal{L}_{\mathrm{SFT}}
= - \sum_{t=1}^{T} \log p_{\theta}\left(y^{*}_{t} \mid y^{*}_{<t}, t, v \right),
\label{eq:sft}
\end{equation}
\vspace{-0.5mm}
where $\theta$ denotes model parameters. 
We show in ablation experiments (Ablation-II (Section~\ref{sec:ablation})) that this cold-start stage is crucial for endowing the model with essential detection and explanation abilities. 
Without this initialization, the base model’s forgery recognition capacity remains weak, leading to sparse rewards in the subsequent RL stage and preventing effective learning of meaningful forensic cues.

\subsection{RL Training}
\label{sec:rl}
We observe that during the data annotation process, human annotators struggled to identify precise artifact cues in certain high-quality generated samples. 
Conversely, for low-quality samples, human-provided labels often contain significant noise. 
Therefore, in this stage, we employ reinforcement learning (RL) to elicit the model's inherent capability for self-coherent forgery cue discovery. 
This approach also aims to continuously improve adaptability to new domains, mitigating the need for iterative manual annotation.  
We adopt Group Relative Policy Optimization (GRPO)~\cite{shao2024deepseekmath,guo2025deepseek} as our RL algorithm, where we re-design the reward score to adapt it to our task.

For each query-completion pair $(x, y)$, the total reward $R(x, y)$ is defined as:
\begin{equation}
\label{eq:reward_total}
R(x,y) = w_{acc} \cdot r_{acc}(x, y) + w_{chk} \cdot r_{chk}(x, y)
\end{equation}
where $w_{acc}=0.8$ and $w_{chk}=0.2$ in our experiments. The rewards $r_{acc}(x,y)$ and $r_{chk}(x, y)$ are defined as follows:

\noindent\textbf{Accuracy Reward $r_{acc}(x, y)$.} We apply an asymmetric reward structure with more severe penalties for false positives: 
\begin{equation}
\label{eq:reward_acc}
r_{acc}(x, y) = \begin{cases}
1.0 & \text{if } y_{pred} = y_{gt}\\
0.0 & \text{if } y_{gt} = \text{``Fake''} \land y_{pred} = \text{``Real''} \\
-0.2 & \text{if } y_{gt} = \text{``Real''} \land y_{pred} = \text{``Fake''} \\
\end{cases}
\end{equation}

\noindent\textbf{Check Reward $r_{chk}(x,y)$.} This reward is activated only when the model's response adheres to the prescribed format $F_{\text{outer}}$. We extract the number of valid check blocks $N_{\text{check}}$ from the model output using regular expressions. The matching pattern follows $F_{\text{fake}}$ when the prediction is ``Fake'', and $F_{\text{real}}$ when the prediction is ``Real'':
\begin{equation}
\label{eq:reward_chk}
r_{chk}(x,y)=min(ln(1+N_{check}),ln(1+3))
\end{equation}

Our reward function encourages active cue exploration while strictly supervising the classification. We observed that symmetric penalties for both error types (false positives and false negatives) caused the model to overfit and develop a strong bias towards predicting ``Fake''. This stems from the inherent asymmetry of the task, i.e., identifying ``Fake'' requires finding just one artifact, while confirming ``Real'' requires comprehensively ruling out all inconsistencies. We validate our asymmetric design in Ablation-III (Section~\ref{sec:ablation}).

%% file: tables/main_results.tex
\begin{table*}[!t]
  \centering
\caption{Detection performances on ViF-Bench.}
    \vspace{-3mm}
  \resizebox{\textwidth}{!}{%
  \addtolength{\tabcolsep}{-3pt}
    \begin{tabular}{
    c|c|cccccc|ccccccc|cccccc|c
    }
      \toprule
        \multirow{2}{*}{Method} &
        \multirow{2}{*}{\makecell{Metric}} &
        
        \makecell{Wan2.1\\-1.3B} &
        \multicolumn{1}{c|}{\makecell{CogV\\-X1.5}} &
        
        \multicolumn{2}{c|}{\makecell{Wan2.2\\-5B}} &
        \multicolumn{2}{c|}{\makecell{Hunyuan\\Video}} &
        
        \makecell{VACE\\-1.3B} &
        
        \multicolumn{2}{c|}{\makecell{Wan2.2\\-14B}} &
        \multicolumn{2}{c|}{\makecell{Skyreels\\-V2}} &
        \multicolumn{2}{c|}{\makecell{LTX-Video\\-13B}} &
        
        \makecell{Gen4\\-Turbo} &
        \makecell{Hai-\\luo-02} &
        \makecell{Pika\\-V2} &
        \makecell{Pixverse\\-V4-5} &
        \makecell{Kling\\-V1} &
        Sora-2 &
        Mean
        \\
        
        \cline{3-22}
        
         &  &
        
        \multicolumn{2}{c|}{T2V} &
              T2V & I2V &
              T2V & I2V &
              T2V & 
              T2V & I2V &
              T2V & I2V &
              T2V & I2V &
              \multicolumn{6}{c|}{T2V} & \textbackslash
        \\
        \midrule
        \multicolumn{22}{c}{\textbf{Binary Detectors}} \\
        \midrule
         
      \multirow{3}{*}{AIGVDet} 
        & Acc
        & 82.12 & 81.21 & 67.58 & 62.73 & 77.44 & 90.00 & 58.48 & 71.52 & 72.12 & 75.15 & 66.46 & 65.31 & 67.48 & 56.25 & 70.07 & 55.30 & 73.68 & 58.87 & 60.67 & 69.08
          \\
        & R
        & 70.91 & 69.09 & 41.82 & 32.12 & 61.59 & \underline{96.92} & 23.64 & 49.70 & 50.91 & 56.97 & 39.63 & 37.50 & 41.72 & 17.86 & 43.80 & 15.89 & 53.95 & 21.99 & 26.67 & 44.88
          \\
        & F1
        & 79.86 & 78.62 & 56.33 & 46.29 & 73.19 & 90.65 & 36.28 & 63.57 & 64.62 & 69.63 & 54.17 & 51.95 & 56.20 & 28.99 & 59.41 & 26.23 & 67.21 & 34.83 & 40.40 & 56.76  
         
          \\
       \hline
      \multirow{3}{*}{DeMamba} 
        & Acc
        & 65.45 & 65.76 & 65.76 & 60.00 & 65.85 & 60.30 & 61.82 & 65.76 & 61.82 & 65.76 & 64.02 & 64.06 & 64.11 & 62.50 & 64.23 & 66.89 & 66.78 & 67.02 & 63.67 & 64.29
          \\
        & R
        & \underline{99.39} & \textbf{100.00} & \textbf{100.00} & \underline{88.48} &  \textbf{100.00} & 89.09 & \underline{92.12} & \textbf{100.00} & \underline{92.12} & \textbf{100.00} & \underline{96.34} & 96.88 & \underline{96.32} & \underline{91.96} & 97.81 & \textbf{100.00} & \textbf{100.00} & \textbf{100.00} & \underline{96.00} & \underline{96.66}
          \\
        & F1
        & 74.21 & 74.49 & 74.49 & 68.87 & 74.55 & 69.18 & 70.70 & 74.49 & 70.70 & 74.49 & 72.81 & 72.94 & 72.85 & 71.03 & 73.22 & 75.12 & 75.06 & 75.20 & 72.54 & 73.00
         
          \\
       \hline

      \multirow{3}{*}{NSG-VD} 
        & Acc
        & 50.00 & 49.39 & 50.30 & 49.70 & 50.30 & 49.70 & 50.00 & 50.00 & 49.39 & 49.70 & 49.09 & 49.06 & 50.00 & 50.45 & 50.36 & 48.34 & 48.68 & 49.29 & 49.67 & 49.65
          \\
        & R
        & \underline{99.39} & 98.18 & \textbf{100.00} & \textbf{98.79} &  \textbf{100.00} & \textbf{98.79} & \textbf{99.39} & \underline{99.39} & \textbf{98.18} & \underline{98.79} & \textbf{97.56} & \underline{97.50} & \textbf{99.39} & \textbf{100.00} & \textbf{100.00} & 96.03 & 96.71 & \underline{97.87} & \textbf{98.67} & \textbf{98.66}
          \\
        & F1
        & 66.53 & 65.99 & 66.80 & 66.26 & 66.80 & 66.26 & 66.53 & 66.53 & 65.99 & 66.26 & 65.71 & 65.68 & 66.53 & 66.87 & 66.83 & 65.02 & 65.33 & 65.87 & 66.22 & 66.21
         
          \\
       \midrule
      \multicolumn{22}{c}{\textbf{Open-source Multimodal Large Language Models}} \\
      \midrule

      \multirow{3}{*}{\makecell{Video-\\LLaMa-3(7B)}} 
        & Acc
        & 50.92 & 53.68 & 51.84 & 51.23 & 50.62 & 51.53 & 52.45 & 50.00 & 50.31 & 50.31 & 50.00 & 50.31 & 52.45 & 50.90 & 50.37 & 49.67 & 49.67 & 50.00 & 51.01 & 50.91
         
          \\
        & R 
        & 2.45 & 7.98 & 4.29 & 3.07 & 1.85 & 3.68 & 5.52 & 0.61 & 1.23 & 1.23 & 0.62 & 1.25 & 5.52 & 2.70 & 0.74 & 0.00 & 0.00 & 0.71 & 2.03 & 2.39
         
          \\
        & F1
        & 4.76 & 14.69 & 8.19 & 5.92 & 3.61 & 7.06 & 10.40 & 1.21 & 2.41 & 2.41 & 1.22 & 2.45 & 10.40 & 5.22 & 1.46 & 0.00 & 0.00 & 1.41 & 3.97 & 4.57
         
          \\
       \hline
      
      \multirow{3}{*}{\makecell{Qwen-\\2.5-VL(3B)}} 
        & Acc
        & 52.12 & 54.85 & 50.61 & 50.00 & 50.00 & 50.00 & 50.91 & 50.61 & 50.91 & 50.00 & 49.7 & 52.5 & 49.69 & 50.00 & 50.36 & 50.99 & 50.33 & 49.65 & 50.00 & 50.7
         
          \\
        & R 
        & 5.45 & 10.91 & 2.42 & 1.21 & 1.22 & 1.54 & 3.03 & 2.42 & 3.03 & 1.21 & 0.61 & 6.25 & 0.61 & 1.79 & 1.46 & 3.31 & 1.97 & 0.00 & 1.33 & 2.62
         
          \\
        & F1
        & 10.23 & 19.46 & 4.68 & 2.37 & 2.38 & 2.99 & 5.81 & 4.68 & 5.81 & 2.37 & 1.20 & 11.63 & 1.20 & 3.45 & 2.86 & 6.33 & 3.82 & 0.00 & 2.60 & 4.94
         
          \\
       \hline
      \multirow{3}{*}{\makecell{Qwen-\\2.5-VL(7B)}} 
        & Acc
        & 51.21 & 50.3 & 49.7 & 51.21 & 51.22 & 51.21 & 50.91 & 50.3 & 50.61 & 49.39 & 50.00 & 50.62 & 51.53 & 50.00 & 50.00 & 50.99 & 50.00 & 49.65 & 52.00 & 50.57
         
          \\
        & R
        & 4.85 & 3.03 & 1.82 & 4.85 & 4.88 & 4.85 & 4.24 & 3.03 & 3.64 & 1.21 & 2.44 & 3.75 & 5.52 & 2.68 & 2.19 & 3.97 & 2.63 & 1.42 & 6.00 & 3.53
         
          \\
        & F1
        & 9.04 & 5.75 & 3.49 & 9.04 & 9.09 & 9.04 & 7.95 & 5.75 & 6.86 & 2.34 & 4.65 & 7.06 & 10.23 & 5.08 & 4.20 & 7.50 & 5.00 & 2.74 & 11.11 & 6.63
        
          \\
       \hline
      \multirow{3}{*}{\makecell{Qwen-\\2.5-VL(72B)}} 
        & Acc
        & 52.42 & 51.82 & 50.91 & 50.3 & 50.61 & 51.52 & 54.55 & 51.21 & 49.39 & 50.00 & 50.30 & 53.12 & 52.45 & 50.00 & 51.82 & 53.31 & 51.97 & 48.58 & 49.67 & 51.26
         
          \\
        & R
        & 9.09 & 7.88 & 6.06 & 4.85 & 5.49 & 7.27 & 13.33 & 6.67 & 3.03 & 4.24 & 4.88 & 10.62 & 9.20 & 6.25 & 8.76 & 10.60 & 8.55 & 2.13 & 4.00 & 6.99
          \\
        & F1
        & 16.04 & 14.05 & 10.99 & 8.89 & 10.00 & 13.04 & 22.68 & 12.02 & 5.65 & 7.82 & 8.94 & 18.48 & 16.22 & 11.11 & 15.38 & 18.50 & 15.12 & 3.97 & 7.36 & 12.44
          \\
       \hline
      \multirow{3}{*}{\makecell{Intern-\\VL-3(8B)}} 
        & Acc
        & 48.16 & 57.98 & 46.01 & 48.77 & 50.00 & 50.61 & 55.83 & 45.71 & 46.01 & 45.40 & 46.60 & 50.62 & 49.69 & 45.95 & 47.06 & 50.00 & 48.00 & 44.29 & 46.28 & 48.58
         
          \\
        & R
        & 13.50 & 33.13 & 9.20 & 14.72 & 17.28 & 18.40 & 28.83 & 8.59 & 9.20 & 7.98 & 10.49 & 18.12 & 16.56 & 11.71 & 10.29 & 17.33 & 11.33 & 5.71 & 9.46 & 14.31
          \\
        & F1
        & 20.66 & 44.08 & 14.56 & 22.33 & 25.69 & 27.15 & 39.50 & 13.66 & 14.56 & 12.75 & 16.43 & 26.85 & 24.77 & 17.81 & 16.28 & 25.74 & 17.89 & 9.30 & 14.97 & 21.31
          \\
        \midrule
      \multicolumn{22}{c}{\textbf{Proprietary Multimodal Large Language Models}} \\
      \midrule
      
      \multirow{3}{*}{GPT-4.1-mini} 
        & Acc
        & 55.83 & 59.82 & 53.37 & 57.06 & 55.25 & 54.60 & 54.60 & 53.68 & 53.68 & 51.23 & 54.94 & 54.69 & 55.83 & 53.60 & 50.00 & 54.33 & 55.33 & 50.71 & 48.99 & 54.08
         
          \\
        & R
        & 18.40 & 26.38 & 13.50 & 20.86 & 17.28 & 15.95 & 15.95 & 14.11 & 14.11 & 9.20 & 16.67 & 16.25 & 18.40 & 12.61 & 8.09 & 16.00 & 16.67 & 8.57 & 4.05 & 14.90
         
          \\
        & F1
        & 29.41 & 39.63 & 22.45 & 32.69 & 27.86 & 26.00 & 26.00 & 23.35 & 23.35 & 15.87 & 27.00 & 26.40 & 29.41 & 21.37 & 13.92 & 25.95 & 27.17 & 14.81 & 7.36 & 24.21
         
          \\
       \hline
      \multirow{3}{*}{\makecell{Gemini\\-2.5-flash}} 
        & Acc
        & 57.67 & 52.76 & 57.67 & 55.21 & 55.56 & 50.92 & 58.90 & 49.08 & 49.39 & 53.07 & 46.30 & 57.19 & 59.20 & 53.60 & 50.74 & 60.33 & 57.33 & 45.36 & 43.58 & 53.36
         
          \\
        & R
        & 72.39 & 62.58 & 72.39 & 67.48 & 67.90 & 58.90 & 74.85 & 55.21 & 55.83 & 63.19 & 49.38 & 70.62 & 75.46 & 65.77 & 58.09 & 78.67 & 70.67 & 47.14 & 45.27 & 63.78
         
          \\
        & F1
        & 63.10 & 56.98 & 63.10 & 60.11 & 60.44 & 54.55 & 64.55 & 52.02 & 52.45 & 57.38 & 47.90 & 62.26 & 64.91 & 58.63 & 54.11 & 66.48 & 62.35 & 46.32 & 44.52 & 57.48
         
          \\

\midrule
      \multicolumn{22}{c}{\textbf{MLLM-based Detectors}} \\
      \midrule
      \multirow{3}{*}{BusterX++(7B)} 
        & Acc
        & 54.85 & 59.39 & 52.42 & 50.30 & 59.15 & 49.70 & 50.91 & 62.42 & 49.70 & 65.76 & 50.00 & 56.25 & 50.00 & 50.89 & 61.68 & 76.82 & 75.66 & 52.84 & 52.33 & 56.90
         
          \\
        & R
        & 10.30 & 19.39 & 5.45 & 1.21 & 18.90 & 0.00 & 2.42 & 25.45 & 0.00 & 32.12 & 0.61 & 13.12 & 0.61 & 2.68 & 24.09 & 54.30 & 51.97 & 5.67 & 5.33 & 14.40
         
          \\
        & F1
        & 18.58 & 32.32 & 10.29 & 2.38 & 31.63 & 0.00 & 4.71 & 40.38 & 0.00 & 48.40 & 1.20 & 23.08 & 1.21 & 5.17 & 38.60 & 70.09 & 68.10 & 10.74 & 10.06 & 21.94
          \\
       \hline
      \multirow{3}{*}{\textbf{Ours-SFT(7B)}} 
        & Acc
        & \textbf{97.58} & \textbf{97.58} & \underline{90.30} & \underline{84.24} & \underline{93.9} & \underline{92.12} & \textbf{79.09} & \textbf{96.67} & \underline{83.03} & \underline{95.15} & \underline{76.83} & \textbf{96.25} & \underline{80.67} & \underline{74.55} & \textbf{97.08} & \textbf{97.35} & \textbf{97.37} & \textbf{95.04} & \underline{87.33} & \underline{90.11}
         
          \\
        & R
        & \underline{99.39} & \underline{99.39} & 84.85 & 72.73 & 92.07 & 88.48 & 62.42 & 97.58 & 70.3 & 94.55 & 57.93 & 96.88 & 65.64 & 54.46 & \underline{98.54} & \underline{99.34} & \underline{99.34} & 95.04 & 79.33 & 84.65
         
          \\
        & F1
        & \textbf{97.62} & \textbf{97.62} & \underline{89.74} & \underline{82.19} & \underline{93.79} & \underline{91.82} & \underline{74.91} & \textbf{96.7} & \underline{80.56} & \underline{95.12} & \underline{71.43} & \textbf{96.27} & \underline{77.26} & \underline{68.16} & \textbf{97.12} & \textbf{97.4} & \textbf{97.42} & \textbf{95.04} & 86.23 & \underline{88.76}
         
          \\
          \hline
        \multirow{3}{*}{\textbf{Ours-RL(7B)}} 
        & Acc
        & \underline{96.97} & \underline{96.36} & \textbf{92.12} & \textbf{87.58} & \textbf{94.82} & \textbf{93.64} & \underline{79.09} & \underline{96.36} & \textbf{84.55} & \textbf{95.76} & \textbf{78.96} & \underline{95.94} & \textbf{83.74} & \textbf{79.46} & \underline{95.99} & \underline{96.36} & \underline{96.05} & \underline{94.68} & \textbf{91.00} & \textbf{91.02}
         
          \\
        & R
        & \textbf{100.00} & 98.79 & \underline{90.30} & 81.21 & \underline{95.73} & 93.33 & 64.24 & 98.79 & 75.15 & 97.58 & 64.02 & \textbf{98.12} & 73.62 & 66.07 & \underline{98.54} & \underline{99.34} & 98.68 & 96.45 & 88.67 & 88.35
         
          \\
        & F1
        & \underline{97.06} & \underline{96.45} & \textbf{91.98} & \textbf{86.73} & \textbf{94.86} & \textbf{93.62} & \textbf{75.44} & \underline{96.45} & \textbf{82.94} & \textbf{95.83} & \textbf{75.27} & \underline{96.02} & \textbf{81.91} & \underline{76.29} & \underline{96.09} & \underline{96.46} & \textbf{96.15} & \textbf{94.77} & \underline{90.78} & \textbf{90.27}
         
          \\
      \bottomrule
    \end{tabular}%
    \addtolength{\tabcolsep}{3pt}
  }
  \label{tab:main_results}
    \vspace{-3.5mm}
\end{table*}

%% file: sec/5_Experiments.tex
\input{tables/GenVideo}
\input{tables/robustness_study}

\section{Experiments}
\label{sec:experiments}
\subsection{Experimental Setup}
\noindent\textbf{Implementation Details.} We build upon Qwen2.5-VL-7B-Instruct popular in video-related tasks~\cite{wen2025busterx++,feng2025video,li2025videochat,meng2025open}, trained on 8 NVIDIA H200 GPUs. During training, we uniformly sample 16 frames from each video and resize them to 256p. 
In the SFT stage, we perform full-parameter fine-tuning with batch size 1 per device for 5 epochs at learning rate 1e-5. 
In the RL stage, we set the actor learning rate to 5e-7 and the KL coefficient to 0.02. For binary detectors, we use identical video pairs with fake/real labels and follow their original training protocols.

\input{tables/ablations}

\noindent\textbf{Evaluation Protocols.} For binary detectors, use weights trained on our dataset and follow their preprocessing scripts. 
For off-the-shelf MLLMs (like GPT-4.1-mini~\cite{openai_gpt4.1}) and ours, we apply chain-of-thought prompts (Figure~\ref{fig:method_overview}) that guide step-by-step inspection for AI-generated artifacts. 
For BusterX++, we follow its original prompt~\cite{wen2025busterx++} to align with its evaluation setups. 
We report accuracy, recall, and F1-score on both our benchmark and GenVideo~\cite{chen2024demamba}. 

\subsection{Main Results}
\noindent\textbf{Results on Our Benchmark.} We compare with three baseline groups: (1) Binary detectors including AIGVDet~\cite{bai2024ai}, DeMamba~\cite{chen2024demamba}, and the recent NSG-VD~\cite{zhang2025physics}. (2) Off-the-shelf MLLMs including VideoLLaMA-3~\cite{Video-llama}, Qwen2.5-VL series~\cite{QWen2.5VL}, InternVL-3~\cite{wang2025internvl3}, GPT-4.1-mini~\cite{openai_gpt4.1}, and Gemini-2.5-flash~\cite{google_gemini2.5}. (3) MLM-based detector, i.e., BusterX++~\cite{wen2025busterx++} which is the only open-sourced implementation available. 
Table~\ref{tab:main_results} shows our model consistently outperforms all baselines, achieving \textbf{+26.73\%} absolute accuracy and \textbf{+17.27\%} F1 over the second-best DeMamba. Compared to MLLM baselines, we notably achieve +34.12\% accuracy, +24.57\% recall, and +32\% F1. Our RL training further improves over SFT, especially on hard I2V samples, with +3.74\% on recall. 

\noindent\textbf{Results on GenVideo Benchmark.} GenVideo serves as our out-of-domain test, containing low-quality samples from outdated generators with near-static content~\cite{chen2025genworld}. 
Our model achieves +11.07\% accuracy over the best binary detector and +7.8\% accuracy, +16.9\% recall over Skyra-SFT, as shown in Table~\ref{tab:GenVideo}. 
To demonstrate the ability to quickly adapt to OOD scenarios, we initialize our framework with Skyra-SFT and perform RL training on only 2.2K data randomly selected from the GenVideo-100K training set, following the ``many-to-many'' settings in GenVideo~\cite{chen2024demamba}. The RL training process, without any additional human-annotation, and is trained for only 1 epoch, quickly adapting Shyra to such a new domain, with the resulting model Skyra-RL-GenVideo achieving a \textbf{+19.22\%} accuracy, \textbf{+42.06\%} recall, and \textbf{+31\%} F1-score gain than Skyra-RL.

\noindent\textbf{Robustness Study.}
Real-world videos are usually represented in degraded formats, causing potential perturbations on the detecting performance. We evaluate robustness under five degradation types: Compression (JPEG), Transformation (Zoom), Gaussian Noise, Light Transform ($-/+$), and color-transform ($-/+$). Table~\ref{tab:robustness} shows our model maintains state-of-the-art performance under all degradations. 

\subsection{Ablation Study}
\label{sec:ablation}
We conduct ablation studies to validate the effectiveness of our design, as shown in Table.~\ref{tab:ablations}


\noindent\textbf{Effects of Training Strategies: Both the CoT reasoning process and the RL boost model's detection performance.} In this part, we comprehensively evaluate the effectiveness of our method design by conducting the following ablation studies: (1) Without CoT in answer: we trained our model to answer with ``Real/Fake'' when inquired ``Is the video a real or fake video?''. This naive implementation yields near-random performance. Without careful reasoning, the model cannot learn detection signals. (2) Without cold-start: we train the base model to perform AIGC video detection and explanation by CoT prompting and GRPO-based reinforcement learning using the design in Section.~\ref{sec:rl}. While similar approaches, as introduced by DeepSeek-R1-Zero~\cite{guo2025deepseek}, have proven to be effective in various visual tasks~\cite{su2025thinking,tang2025video}, duplicating its success to our tasks is not naive. The resulting model achieves even worse performance than (1). Considering the base model's inability in AI-generated video detection tasks, purely RL can harly equip the model with sufficient artifacts identifying capability without our cold-start initialization process~\cite{yue2025does}. (3) Without the RL stage: our reinforcement training stage further boosts the detection performance of the supervised finetuned model. 

\noindent\textbf{Effect of Reward Design: Direct real-fake binary reward yields suboptimal performance.} In our reward score, we introduce two special designs: asymmetric accuracy reward and inspection reward. When setting the accuracy reward to 0.0 for both false positive and negative, the model quickly overfitted to ``fake'', with a sharp decline in both accuracy (-14.78\%) and F1-score (-9.62\%). We also observe a performance drop when we replace the inspection reward with the normal format reward that only inspects whether the model's response follows $F_{outer}$.

\begin{figure}[!t]
    \centering
    \includegraphics[width=0.47\textwidth]{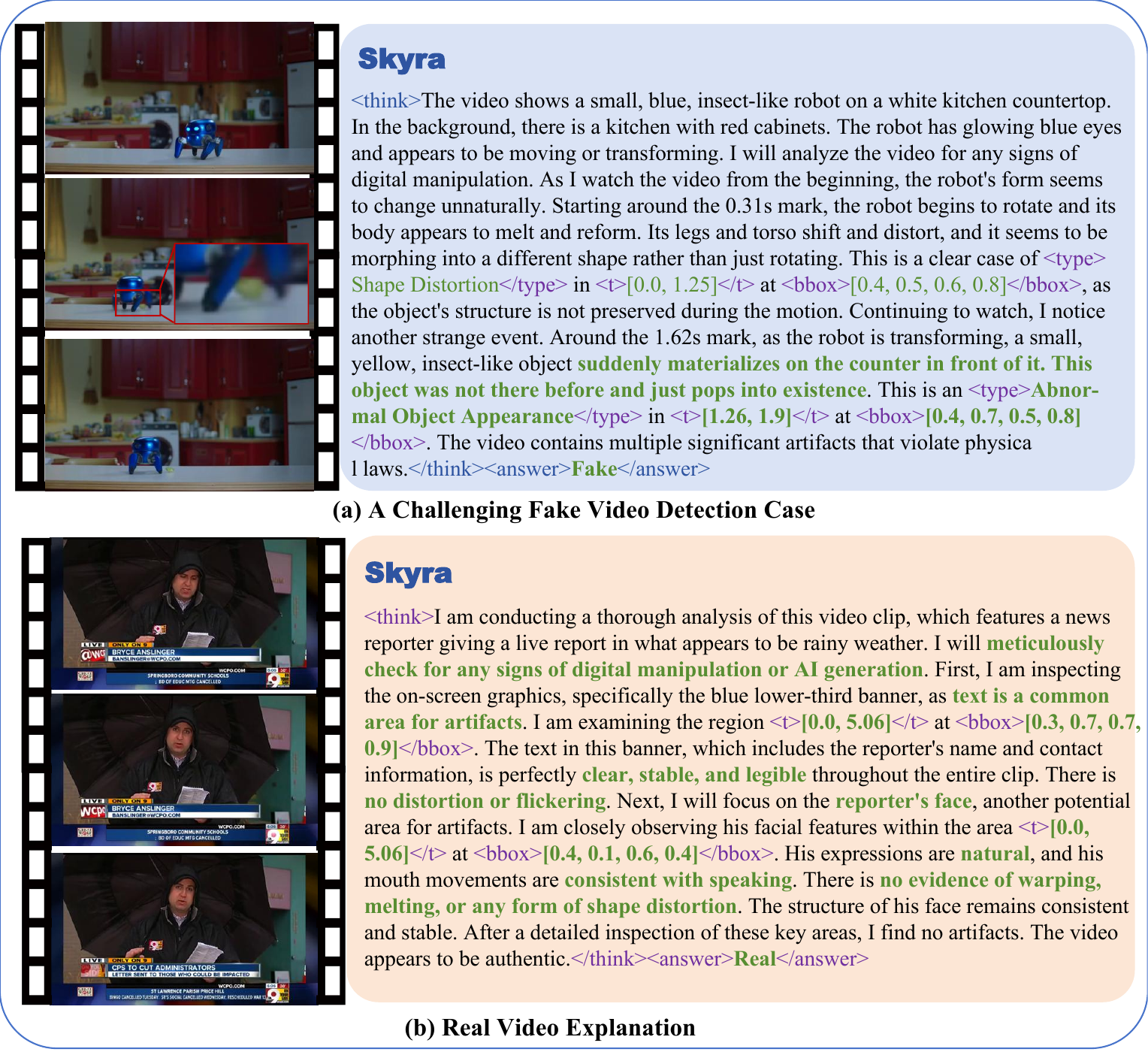} 
    \vspace{-3mm}
    \caption{
        Case Study. More examples are provided in the appendix.
    }
    \vspace{-6mm}
    \label{fig:case_study}
\end{figure}

\subsection{Case Study of the Explanation Ability}
We provide two response cases of Skyra to demonstrate the explanation ability in Figure~\ref{fig:case_study}. Our analysis is as follows: \textbf{(a)} Through our cold-start initialization stage with our high-quality human-annotated dataset, and the RL stage training, which further enhances our model's artifacts perception capability, we are delighted to find that Skyra can perceive tiny-grained AIGC evidence that are even hard for humans to identify. \textbf{(b)} For real videos, Skyra follows a ``description-inspection-review-conclusion'' process, which guides the model's attention to those areas that are likely to contain artifact evidence, avoiding missing any possible forgery evidence, balancing the training process gap between real and AIGC videos, and also making the model's explanation more persuasive. We provide more cases in the Appendix.

%% file: tables/GenVideo.tex
\begin{table*}[t] \small
  \centering

\caption{Detection performances on GenVideo. }
    \vspace{-3mm}
    \begin{tabular}{
    c|c|cccccccccc|c
    }
      \toprule
      Method 
      & 
      Metric & 
      \makecell{Model\\Scope} & 
      \makecell{Morph\\Studio} & 
      \makecell{Moon\\Valley} & 
      \makecell{Hot\\Shot} & 
      \makecell{Show1} & 
      \makecell{Gen2}& 
      \makecell{Crafter} &
      \makecell{Lavie} &
      \makecell{Sora} &
      \makecell{Wild\\Scrape} &
      \makecell{Avg.}
      \\
      
      \midrule
         
         
         

         \multirow{3}{*}{\makecell{AIGVDet}} 
         & ACC
         & 50.36
         & 50.21
         & 50.00
         & 50.07
         & 50.00
         & 50.00
         & 50.18
         & 50.04
         & 50.00
         & 50.00
         & 50.09
         
          \\
         & R
         & 0.71
         & 0.43
         & 0.00
         & 0.14
         & 0.00
         & 0.00
         & 0.36
         & 0.07
         & 0.00
         & 0.00
         & 0.17
         
         \\
         & F1
         & 1.42
         & 0.28
         & 0.00
         & 0.29
         & 0.00
         & 0.00
         & 0.71
         & 0.14
         & 0.00
         & 0.00
         & 0.34
         
         \\
         \hline
       
      \multirow{3}{*}{\makecell{DeMamba}} 
         & ACC
         & \underline{62.00}
         & 61.36
         & 62.06
         & 56.36
         & \underline{61.50}
         & 62.75
         & 61.80
         & 61.29
         & 62.50
         & 55.51
         & 60.71
         
          \\
         & R
         & \underline{96.43}
         & \underline{98.43}
         & \textbf{99.84}
         & \textbf{99.84}
         & \underline{85.57}
         & \underline{97.14}
         & \underline{98.91}
         & \underline{98.86}
         & \underline{97.93}
         & \underline{88.11}
         & \underline{95.94}
         
         \\
         & F1
         & \underline{71.32}
         & \underline{71.81}
         & 72.46
         & 66.22
         & \underline{71.61}
         & \underline{72.65}
         & 72.13
         & 71.67
         & 72.37
         & \underline{66.45}
         & \underline{70.91}
         
         \\
         \hline
      \multirow{3}{*}{\makecell{NSG-VD}} 
        & ACC
         & 49.79
         & 50.14
         & 49.60
         & 50.29
         & 50.07
         & 50.22
         & 49.86
         & 50.29
         & 50.00
         & 40.94
         & 49.12
         \\
        & R
         & \textbf{99.14}
         & \textbf{99.29}
         & \underline{98.88}
         & \underline{99.14}
         & \textbf{99.43}
         & \textbf{99.28}
         & \textbf{99.00}
         & \textbf{99.21}
         & \textbf{100.00}
         & \textbf{99.36}
         & \textbf{99.27}
         \\
        & F1
         & 66.38
         & 66.57
         & 66.24
         & \underline{66.60}
         & 66.57
         & 66.60
         & 66.38
         & 66.62
         & 66.67
         & 57.86
         & 65.65
         \\

       \midrule

      \multirow{3}{*}{\textbf{\makecell{Ours-SFT(7B)}}} 
         & ACC
         & 52.71
         & 62.07
         & 69.01
         & 53.21
         & 51.64
         & 66.99
         & 77.04
         & 67.86
         & 77.86
         & 61.62
         & 63.98
        
          \\
         & R
         & 6.14
         & 24.43
         & 38.98
         & 7.00
         & 4.86
         & 34.64
         & 54.72
         & 36.29
         & 55.36
         & 24.56
         & 28.70
         \\
         & F1
         & 11.50
         & 39.18
         & 55.17
         & 13.01
         & 9.13
         & 51.21
         & 70.44
         & 53.03
         & 71.26
         & 39.02
         & 41.00
         \\
         \hline
        \multirow{3}{*}{\textbf{\makecell{Ours-RL(7B)}}} 
         & ACC
         & 57.21
         & \underline{69.14}
         & \underline{81.71}
         & \underline{57.50}
         & 56.71
         & \underline{75.94}
         & \underline{86.16}
         & \underline{78.50}
         & \underline{86.61}
         & \underline{68.34}
         & \underline{71.78}
        
          \\
         & R
         & 16.71
         & 40.29
         & 65.34
         & 17.57
         & 16.57
         & 54.49
         & 73.75
         & 59.14
         & 73.21
         & 38.88
         & 45.60
         \\
         & F1
         & 28.09
         & 56.63
         & \underline{78.13}
         & 29.25
         & 27.68
         & 69.37
         & \underline{84.20}
         & \underline{73.34}
         & \underline{84.54}
         & 55.11
         & 59.00
         \\
         \hline
         \multirow{3}{*}{\textbf{\makecell{Ours-RL-\\GenVideo(7B)}}} 
         & ACC
         & \textbf{79.93}
         & \textbf{94.43}
         & \textbf{96.09}
         & \textbf{88.50}
         & \textbf{83.50}
         & \textbf{95.25}
         & \textbf{95.92}
         & \textbf{94.32}
         & \textbf{95.54}
         & \textbf{86.56}
         & \textbf{91.00}
        
          \\
         & R
         & 66.29
         & 95.86
         & 97.92
         & 82.71
         & 72.00
         & 96.23
         & 98.07
         & 94.29
         & 95.64
         & 78.63
         & 87.66
         \\
         & F1
         & \textbf{76.76}
         & \textbf{94.51}
         & \textbf{96.16}
         & \textbf{87.79}
         & \textbf{81.36}
         & \textbf{95.30}
         & \textbf{96.01}
         & \textbf{94.32}
         & \textbf{95.50}
         & \textbf{85.41}
         & \textbf{90.00}
         \\
      \bottomrule
    \end{tabular}%
  \label{tab:GenVideo}
    \vspace{-3mm}
\end{table*}

%% file: tables/robustness_study.tex
\begin{table*}[t] \small
  \centering
  
\caption{Robustness evaluation of different detectors on ViF-Bench. }
  
    \vspace{-3mm}
  \setlength{\tabcolsep}{10pt}
    \begin{tabular}{
    c|c|c|ccccccc
    }
      \toprule
      \multirow{2}{*}{Method} 
      & 
      \multirow{2}{*}{Metric} & 
      \multirow{2}{*}{Origin} &
      \multirow{2}{*}{\makecell{Compre-\\ssion}} &
      \multirow{2}{*}{\makecell{Trans-\\formation}} &
      \multirow{2}{*}{\makecell{Gaussian\\Noise}} & 
      \multicolumn{2}{c}{\makecell{Light-transform}} & 
      \multicolumn{2}{c}{\makecell{Color-transform}} 
      \\
      \cline{7-10}
      & & & & & & $(-)$ & $(+)$ & $(-)$ & $(+)$ 
      \\
      \midrule
         
         
         

         \multirow{3}{*}{\makecell{AIGVDet}} 
         & ACC
         & 69.08 & 70.33 & 54.91 & 56.30 & 62.33 & 64.01 & 72.93 & 77.52

          \\
         & R
         & 44.88 & 50.42 & 0.57 & 5.57 & 20.84 & 22.87 & 54.83 & 64.88

          \\
         & F1
         & 56.76 & 58.43 & 1.03 & 8.57 & 30.21 & 33.13 & 62.18 & 69.25

          \\
         \hline
      \multirow{3}{*}{\makecell{DeMamba}} 
         & ACC
         & 64.29 & 63.94 & 64.62 & 63.18 & 64.06 & 62.81 & 63.25 & 63.69

          \\
         & R
         & \underline{96.66} & \underline{96.68} & \underline{96.79} & \underline{96.41} & \underline{96.92} & \underline{96.90} & \underline{96.55} & \underline{96.81}

          \\
         & F1
         & 73.00 & 72.28 & 72.69 & 71.80 & 72.40 & 71.71 & 71.88 & 72.18

        \\
         \hline
      \multirow{3}{*}{\makecell{NSG-VD}} 
         & ACC
         & 49.65 & 48.71 & 48.60 & 48.97 & 48.71 & 49.14 & 48.73 & 49.21

          \\
         & R
         & \textbf{98.66} & \textbf{99.20} & \textbf{98.97} & \textbf{99.72} & \textbf{99.19} & \textbf{98.80} & \textbf{99.22} & \textbf{98.94}

          \\
         & F1
         & 66.21 & 65.34 & 65.24 & 65.58 & 65.34 & 65.44 & 65.35 & 65.50

          \\
       \midrule
      \multirow{3}{*}{\makecell{Qwen-\\2.5-VL(7B)}} 
         & ACC
         & 51.26
         & 49.95
         & 48.93
         & 50.68
         & 48.93
         & 54.90
         & 48.93
         & 54.94
         
          \\
         & R
         & 6.99
         & 4.16
         & 5.12
         & 5.15
         & 5.13
         & 10.41
         & 5.13
         & 10.48
         
         \\
         & F1
         & 12.44
         & 7.60
         & 9.04
         & 9.41
         & 9.04
         & 17.20
         & 9.04
         & 17.31
         
         \\
         \hline
      \multirow{3}{*}{\makecell{Buster-\\X++(7B)}} 
         & ACC
         & 56.90
         & 55.02
         & 59.10
         & 59.54
         & 59.12
         & 54.90
         & 59.12
         & 54.94
         
          \\
         & R
         & 14.40
         & 10.64
         & 23.04
         & 23.27
         & 23.04
         & 10.41
         & 23.04
         & 10.48
         
         \\
         & F1
         & 21.94
         & 17.43
         & 33.09
         & 33.04
         & 33.09
         & 17.20
         & 33.09
         & 17.31
         
         \\
         \hline
      \multirow{3}{*}{\textbf{\makecell{Ours-SFT(7B)}}} 
         & ACC
         & \underline{90.11}
         & \underline{80.52}
         & \textbf{86.21}
         & \textbf{83.70}
         & \textbf{88.67}
         & \underline{88.12}
         & \textbf{88.50}
         & \underline{88.51}
        
          \\
         & R
         & 84.65
         & 85.54
         & 91.31
         & 94.77
         & 89.93
         & 80.68
         & 85.21
         & 80.85
         
         \\
         & F1
         & \underline{88.76}
         & \underline{81.06}
         & \textbf{86.64}
         & \underline{85.28}
         & \textbf{88.51}
         & \underline{86.18}
         & \textbf{87.43}
         & \underline{86.58}
         
         \\
         \hline
        \multirow{3}{*}{\textbf{\makecell{Ours-RL(7B)}}} 
         & ACC
         & \textbf{91.02}
         & \textbf{80.80}
         & \underline{83.26}
         & \underline{83.48}
         & \underline{83.26}
         & \textbf{90.66}
         & \underline{83.26}
         & \textbf{90.67}
        
          \\
         & R
         & 88.35
         & 88.64
         & 96.37
         & 96.34
         & 96.37
         & 85.78
         & 96.37
         & 85.81
         
         \\
         & F1
         & \textbf{90.27}
         & \textbf{81.93}
         & \underline{85.17}
         & \textbf{85.33}
         & \underline{85.17}
         & \textbf{89.62}
         & \underline{85.17}
         & \textbf{89.64}
         
         \\
      \bottomrule
    \end{tabular}%
  \label{tab:robustness}
    \vspace{-3.6mm}
\end{table*}

%% file: tables/ablations.tex
  

\begin{table*}[!t] \small
  \centering
\setlength{\tabcolsep}{4pt}
\caption{Results of ablation studies.}
\vspace{-3mm}
\begin{tabular}{c|c|ccc|ccc}
\toprule
\multirow{3}{*}{No.} & \multirow{3}{*}{Type} 
& \multicolumn{3}{c|}{Ours-SFT} 
& \multicolumn{3}{c}{Ours-RL} \\
\cline{3-8}
& & Acc & R & F1 & Acc & R & F1 \\
\cline{3-8}
& & 90.11 & 84.65 & 88.76 & 91.02 & 88.35 & 90.27 \\
\midrule

\multirow{3}{*}{Ablation-I} 
& wo CoT 
& 54.04 \textcolor{red}{(-36.07)} 
& 9.36 \textcolor{red}{(-75.29)} 
& 16.72 \textcolor{red}{(-72.04)} 
& \textbackslash & \textbackslash & \textbackslash \\

& w/o Cold-Start 
& \textbackslash & \textbackslash & \textbackslash 
& 50.09 \textcolor{red}{(-40.93)} 
& 0.18 \textcolor{red}{(-88.17)} 
& 0.37 \textcolor{red}{(-89.90)} \\

& w/o RL 
& \textbackslash & \textbackslash & \textbackslash 
& 90.11 \textcolor{red}{(-0.91)} 
& 84.65 \textcolor{red}{(-3.70)} 
& 88.76 \textcolor{red}{(-1.51)} \\
\midrule

\multirow{2}{*}{Ablation-II} 
& wo Asymmetric Reward 
& \textbackslash & \textbackslash & \textbackslash 
& 76.24 \textcolor{red}{(-14.78)} 
& 99.07 \textcolor{green}{(+10.72)} 
& 80.65 \textcolor{red}{(-9.62)} \\

& wo Inspection Reward 
& \textbackslash & \textbackslash & \textbackslash 
& 90.05 \textcolor{red}{(-0.97)} 
& 87.55 \textcolor{red}{(-0.80)} 
& 89.30 \textcolor{red}{(-0.97)} \\
\bottomrule
\end{tabular}%
\vspace{-6mm}
\label{tab:ablations}
\end{table*}

%% file: sec/6_Conclusion.tex
\section{Conclusion and Discussions}
\label{sec:conclusion}
In this paper, we introduce Skyra, a specialized multimodal large language model designed for interpretable, artifact-centric AI-generated video detection. Built upon the fine-grained, human-annotated ViF-CoT-4K dataset and a two-stage training pipeline that integrates supervised initialization with reinforcement learning, Skyra exhibits strong spatio-temporal artifact perception and produces coherent, grounded explanations. Extensive experiments on ViF-Bench and GenVideo demonstrate substantial improvements over existing binary and MLLM-based detectors, while also uncovering systematic patterns in generative artifacts and model reasoning behavior. We hope that Skyra, together with our dataset and benchmark, can support future research toward more transparent, robust, and trustworthy AIGC video detection systems, contributing to the broader effort of mitigating societal risks associated with synthetic media. 

\noindent\textbf{Limitations.} Our training data are still bound by the specific generators and collection pipeline used in ViF-CoT-4K and ViF-Bench. Although we cover a diverse set of recent text-to-video and image-to-video models, the benchmark does not yet encompass all emerging media distributions (e.g., ultra-long videos or non-photorealistic, stylized content). 
Also, Skyra does not assess the intent, context, or potential societal harm of a video. Its natural-language rationales are designed to be persuasive and human-readable, but they may still be overconfident or partially hallucinated. This highlights the importance of calibrated uncertainty estimation, human-in-the-loop use, and complementary safeguards when deploying such models in safety-critical scenarios.

%% file: sec/7_Acknowledge.tex
\section*{Acknowledgement}

This work was supported in part by the National Natural Science Foundation of China under Grant 62441616, Grant 62336004, Grant 62125603, Grant 62306031, Grant 62506198, in part by the China Postdoctoral Science Foundation under Grant 2024M761674.

%% file: sec/X_suppl.tex
\clearpage
\section*{\Large Appendix}
\renewcommand{\thesection}{\Alph{section}}  
\setcounter{section}{0}

\textbf{Content of Appendices}\\
\textbf{Section}~\ref{sec:the_vif_dataset}. The ViF Dataset.\\
\hspace*{1.5em}• \textbf{\S}~\ref{subsec:definition}. Definition of Each Artifact Category.\\
\hspace*{1.5em}• \textbf{\S}~\ref{subsec:annotation_platform}. Annotation Platform.\\
\hspace*{1.5em}• \textbf{\S}~\ref{subsec:cot_prompt}. Chain-of-Though Annotation Prompt Design.\\
\hspace*{1.5em}• \textbf{\S}~\ref{subsec:annotation_details}. Annotation Details, Dataset Statistics, and Training Settings.\\
\hspace*{1.5em}• \textbf{\S}~\ref{subsec:detailed_statistics}. Detailed Statistics of ViF-CoT-4K.\\
\hspace*{1.5em}• \textbf{\S}~\ref{subsec:aigc_examples}. Generated Video Examples.\\
\textbf{Section}~\ref{sec:generalization_ablation}. Analysis of Generalization Ablations.\\
\textbf{Section}~\ref{sec:artifact_cue_analysis}. Analysis of Artifact Cues.\\
\textbf{Section}~\ref{sec:analysis_of_baselines}. Analysis of Baselines' Detection Capabilities.\\
\hspace*{1.5em}• \textbf{\S}~\ref{subsec:binary_detectors}. Binary Detectors.\\
\hspace*{1.5em}• \textbf{\S}~\ref{subsec:off-the-self-MLLM}. Off-the-Shelf MLLMs.\\
\hspace*{1.5em}• \textbf{\S}~\ref{subsec:existing_mllm_detectors}. Existing MLLM-based Detectors.\\
\textbf{Section}~\ref{sec:additional_examples}. Additional Examples.\\
\hspace*{1.5em}• \textbf{\S}~\ref{subsec:design_of_prompt}. Design of Prompts.\\
\hspace*{1.5em}• \textbf{\S}~\ref{subsec:examples_of_skyra}. Examples of Skyra's Responses.\\
\textbf{Section}~\ref{subsec:broader_impacts}. Broader Impacts.\\
\textbf{Section}~\ref{sec:license}. License.\\

\section{The ViF Dataset}
\label{sec:the_vif_dataset}
\subsection{Definition of Each Artifact Category}
\label{subsec:definition}
We provide detailed definitions of each category in our artifact taxonomy (Section~\ref{subsec:taxonomy}) as follows.

\noindent\textbf{Low-Level Forgery.}
This group summarizes characteristic visual cues that frequently make current AI-generated videos appear ``unnatural''. 
These cues typically do not explicitly violate physical laws, but reflect systematic limitations of mainstream video generation models.

\begin{itemize}
  \item \textbf{Texture Anomaly.}
  This category focuses on abnormal patterns in local textures.
  \begin{itemize}
    \item \textit{Structure Anomaly.}
    Regions with rich structures (e.g., fences, grids, lattices) exhibit unnatural distortion, twisting, or interlacing, leading to inconsistent or implausible geometric patterns.
    \item \textit{Texture Jittering.}
    Surface textures show high-frequency flickering or drifting over time, manifesting as crawling patterns, grid-like noise, or temporally unstable blur, instead of stable, physically plausible textures.
    \item \textit{Unnatural Blur.}
    Blur and degradation patterns differ from typical natural degradations, such as Gaussian blur or compression artifacts.
    The blur may be spatially inconsistent, texture-selective, or temporally unstable in a way rarely observed in real videos.
  \end{itemize}

  \item \textbf{Color and Lighting Anomaly.}
  This category captures implausible color or illumination patterns that deviate from natural imaging conditions.
  \begin{itemize}
    \item \textit{Color Over-saturation.}
    Certain regions exhibit excessively saturated or overly vivid colors (often in blue, red, or green), with insufficient tonal variation or shading, making the area visually stand out unnaturally.
    \item \textit{Lighting Inconsistency.}
    Global or local illumination changes abruptly or violently over time, or shows strong intensity fluctuations that cannot be explained by realistic changes of light sources, exposure, or scene configuration.
  \end{itemize}

  \item \textbf{Motion Forgery.}
  This category describes unnatural camera-related motion patterns.
  \begin{itemize}
    \item \textit{Camera Motion Inconsistency.}
    The apparent camera motion is abnormal, such as erratic zooming in/out, unnatural high-frequency panning, or irregular shaking.
    These artifacts are often accompanied by inconsistent changes in object positions, scales, or spatial relations that do not match a physically plausible camera trajectory.
  \end{itemize}
\end{itemize}

\noindent\textbf{Violation of Laws.}
This group contains artifacts that clearly violate real-world constraints, including object permanence, physical laws, causality, and basic common sense.
Detecting these cues generally requires spatio-temporal reasoning and background knowledge about how objects and scenes behave in reality.

\begin{itemize}
  \item \textbf{Object Inconsistency.}
  This category focuses on violations of object permanence and identity over time.
  \begin{itemize}
    \item \textit{Abnormal Object Disappearance.}
    An object disappears suddenly during its motion without any plausible interaction or occlusion.
    For example, a runner on a track vanishes abruptly while continuing to move forward.
    \item \textit{Abnormal Object Appearance.}
    An object suddenly appears and starts to move without any reasonable cause or prior indication.
    For example, a new runner appears out of nowhere on the track in the middle of the video.
    \item \textit{Person Identity Inconsistency.}
    The identity of a person changes over time, especially in facial features or other stable identity cues.
    For example, a person’s face disappears and reappears with clearly different facial characteristics, leading to a mismatch in perceived identity.
    \item \textit{General Object Identity Inconsistency.}
    The identity of a generic object changes significantly over time without any obvious external cause.
    For example, a chair being rotated by a person ends up with a drastically different color or structure compared to its initial state.
    \item \textit{Shape Distortion.}
    Rigid objects exhibit non-rigid deformations during motion.
    For instance, a human body suddenly scales up and down or undergoes frequent surface distortions and twisting that are incompatible with rigid-body motion.
  \end{itemize}

  \item \textbf{Interaction Inconsistency.}
  This category captures physically implausible interactions between multiple objects.
  \begin{itemize}
    \item \textit{Abnormal Rigid-Body Crossing.}
    Rigid objects that should remain non-interpenetrating instead of intersecting or passing through each other.
    For example, a barbell that should move in front of a person’s body passes unrealistically through the person’s head.
    \item \textit{Abnormal Multi-Object Merging.}
    Two or more distinct objects gradually or abruptly merge into a single object during motion, without any plausible explanation (e.g., three people in motion merge into two).
    \item \textit{Abnormal Object Splitting.}
    A single object splits into multiple distinct objects during motion, again without any reasonable cause (e.g., one person splits into two separate people).
    \item \textit{General Interaction Anomaly.}
    Other abnormal or implausible phenomena occurring during interactions between two or more objects, such as missing collisions, inconsistent contact, or contradictory occlusion relations.
  \end{itemize}

  \item \textbf{Unnatural Movement.}
  This category denotes motion patterns that contradict the typical kinematics of humans, animals, or objects.
  \begin{itemize}
    \item \textit{Unnatural Human Movement.}
    Human body motion deviates from normal biomechanics or everyday experience.
    For example, a person walks without leg crossing, exhibiting pure lateral sliding of the legs instead of realistic gait cycles.
    \item \textit{Unnatural Animal Movement.}
    Animal motion is incompatible with known locomotion patterns.
    For example, a running horse moves its hind legs in parallel translation without proper alternating strides.
    \item \textit{Unnatural General Object Movement.}
    Objects other than humans and animals follow trajectories or undergo transformations that are inconsistent with real-world dynamics, such as erratic acceleration, unnatural smoothness, or implausible temporal discontinuities.
  \end{itemize}

  \item \textbf{Violation of Causality Law.}
  This category collects artifacts that violate physical laws or general causal relationships.
  \begin{itemize}
    \item \textit{Violation of Physical Laws.}
    The motion of objects contradicts basic physical principles, such as force–acceleration relationships or conservation laws.
    For example, a ball moves or changes velocity in the absence of any visible force, or instantaneously teleports at unrealistic speeds.
    \item \textit{Violation of General Causality Violation.}
    Events occur without observable causes, or actions fail to produce their expected effects.
    For example, a boy spills milk onto a table, but no milk traces appear on the table surface.
  \end{itemize}

  \item \textbf{Violation of Common Sense.}
  This category covers structural or semantic inconsistencies that conflict with basic commonsense knowledge.
  \begin{itemize}
    \item \textit{Abnormal Human Body Structure.}
    The generated human body deviates from normal anatomical structure.
    Examples include extra or missing body parts (e.g., two heads, three or six fingers), or impossible body bending that is incompatible with human physiology.
    \item \textit{Abnormal General Object Structure.}
    Non-human objects exhibit structures that are inconsistent with their typical shapes or assembly, such as missing essential components or impossible connections.
    \item \textit{Text Distortion.}
    Text appearing in the scene is severely distorted, malformed, or rendered as illegible gibberish without coherent semantic content, beyond mild degradation commonly observed in real footage.
  \end{itemize}
\end{itemize}

\subsection{Annotation Platform}
\label{subsec:annotation_platform}
\begin{figure}[h]
    \centering
    \includegraphics[width=0.45\textwidth]{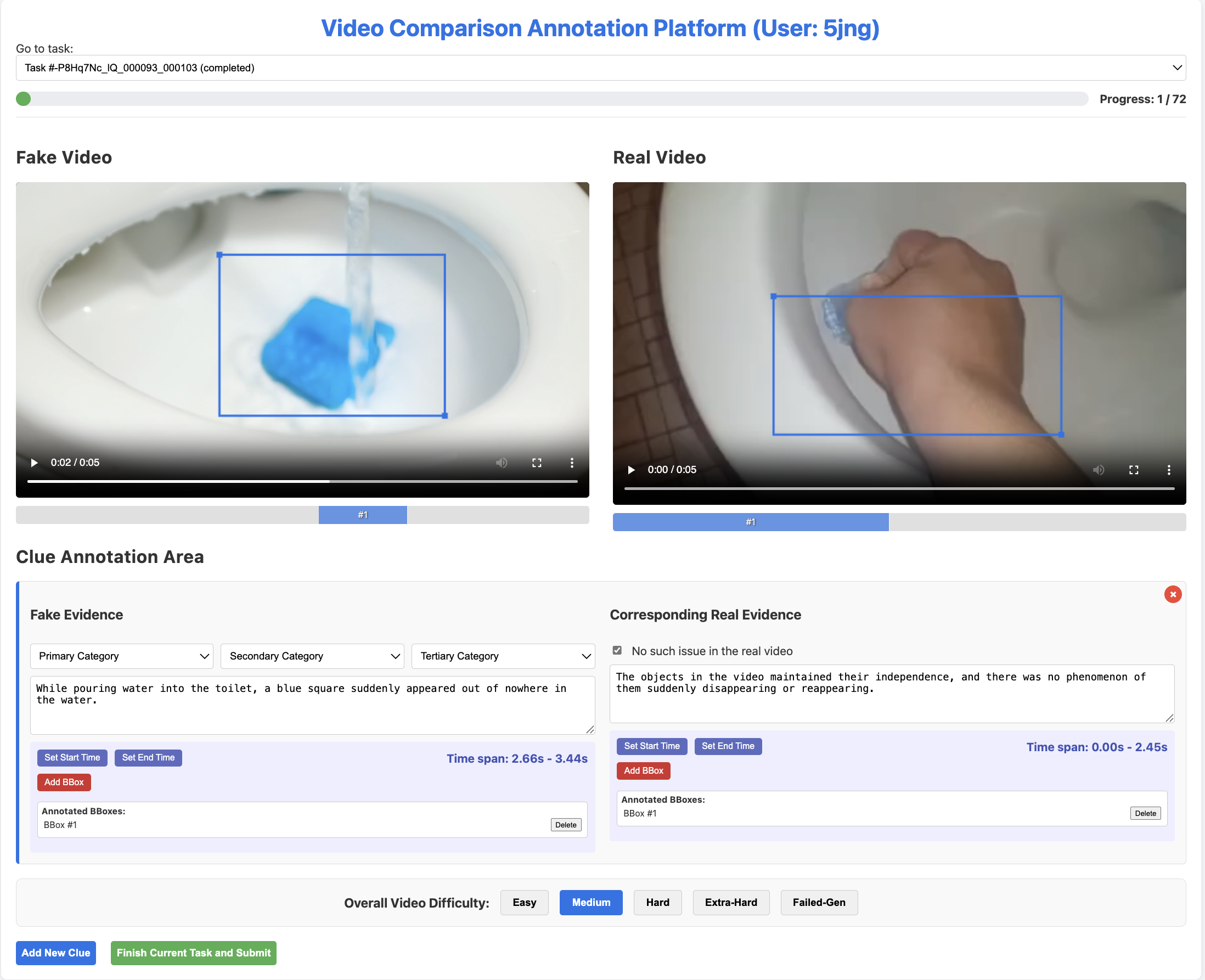}
    \vspace{-3.5mm}
    \caption{
        Annotation platform UI.
    }
    \vspace{-5mm}
    \label{fig:annotation_platform}
\end{figure}

Our annotation platform presents each AI-generated video alongside its corresponding real counterpart in a synchronized comparison view (Figure~\ref{fig:annotation_platform}). This side-by-side layout allows annotators to directly contrast suspicious regions in the fake video with how the same scene should plausibly appear in real footage, making it easier to distinguish genuine physical phenomena from artifacts that only occur in AIGC videos. For every identified clue, annotators are required to select a fine-grained artifact category, provide detailed textual explanations for both the fake and real videos, and supply precise spatio-temporal annotations by marking time spans and bounding boxes in both streams. By enforcing mirrored annotations on fake–real pairs, the platform encourages annotators to explicitly encode both “what is wrong” in the generated video and “what is normal” in the real video, guiding the model toward learning an unbiased perceptual representation that treats real and synthetic content in a symmetric manner.

\subsection{Chain-of-Thought Annotation Prompt Design}
\label{subsec:cot_prompt}
\begin{figure*}[h]
    \centering
    \includegraphics[width=1\textwidth]{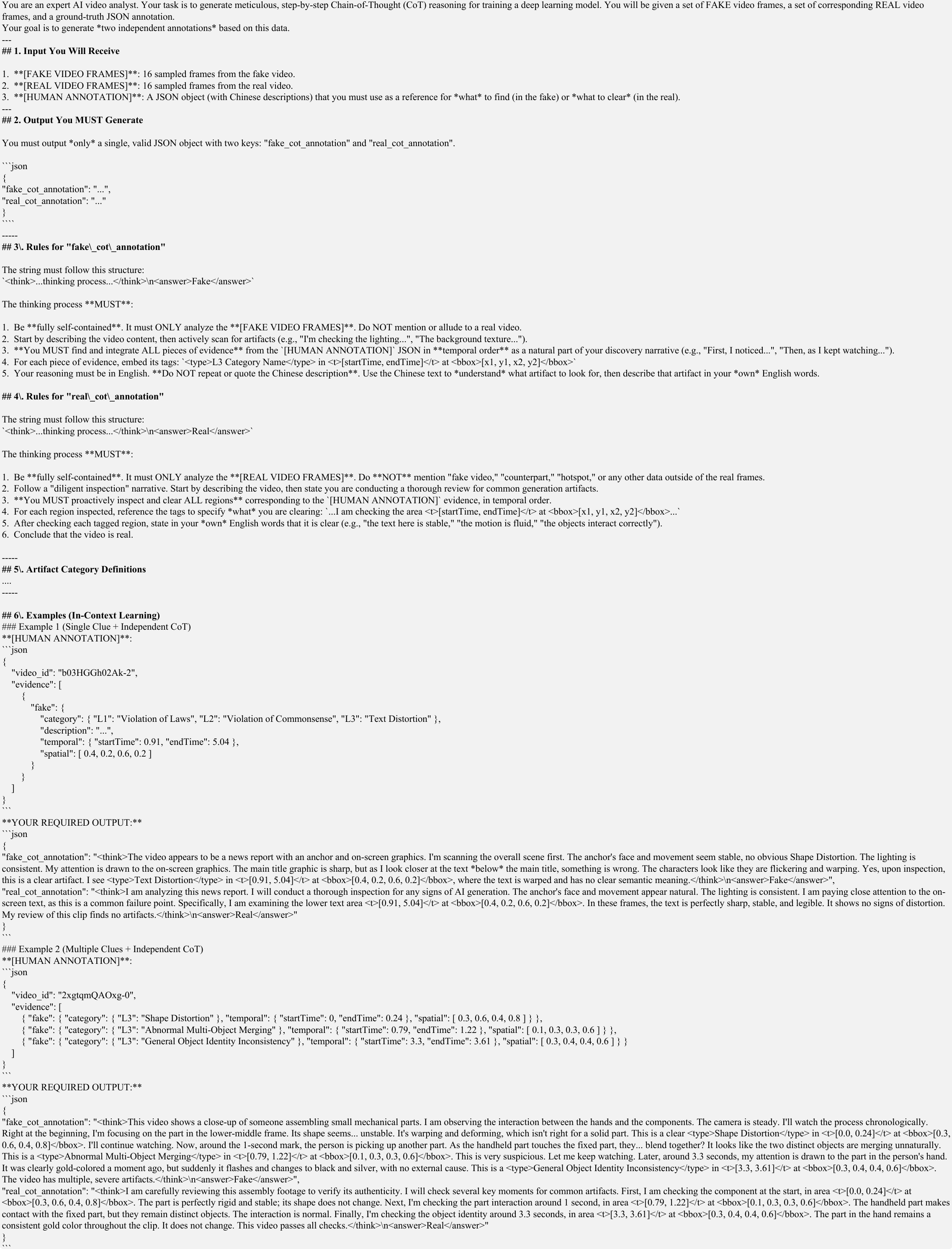}
    \vspace{-7.5mm}
    \caption{
        Chain-of-Thought Annotation Prompt.
    }
    \vspace{-5mm}
    \label{fig:annotation_prompt}
\end{figure*}

To transform concise human annotations into training-ready step-by-step supervision, we design a structured prompt for Gemini-2.5-Pro that operates on each fake–real video pair. For every annotated instance, the model receives sampled frames from the fake and real videos together with the curated artifact \textbf{Type}, \textbf{Textual Explanation}, \textbf{Timestamps}, and \textbf{Bounding Boxes}, and is instructed to produce two independent CoT strings: one that carefully discovers all artifacts in the fake video and one that systematically clears the corresponding regions in the real video. The prompt enforces a standardized JSON output format (with separate \texttt{fake\_cot\_annotation} and \texttt{real\_cot\_annotation} fields), requires explicit tagging of temporal spans and spatial regions, and guides the reasoning process through an \textit{observe–understand–draft–review–conclude} workflow with in-context examples. This design allows us to automatically expand precise but terse human labels into rich, consistent CoT supervision suitable for SFT. The complete prompt is provided in Figure~\ref{fig:annotation_prompt}

\subsection{Annotation Details, Dataset Statistics, and Training Settings}
\label{subsec:annotation_details}
We provide a comprehensive summary of dataset and training specifications in Table~\ref{tab:dataset_training_specs}.

\noindent\textbf{Annotation Protocol.}
A total of 25 trained professional annotators participated in the annotation process. Each annotator followed detailed guidelines covering artifact definitions (Section~\ref{subsec:definition}), annotation platform usage (Section~\ref{subsec:annotation_platform}), and quality standards. The annotation workflow proceeded as follows: (1) annotators watched each AI-generated video alongside its real counterpart in the synchronized comparison view (Figure~\ref{fig:annotation_platform}); (2) for each identified artifact, they selected a fine-grained category from our taxonomy, provided textual explanations, and marked precise temporal spans and bounding boxes; (3) annotations underwent multiple review cycles to ensure quality and inter-annotator consistency.

\noindent\textbf{Video Filtering.}
To ensure the quality and relevance of the collected videos, we apply a two-stage filtering pipeline: (i) \textit{automatic semantic consistency checking} via GPT-4o-mini~\cite{openai2024gpt4omini} between the AI-generated and real videos, which removes pairs with significant semantic drift; and (ii) \textit{manual filtering} during annotation, where annotators remove static, ambiguous, or severely distorted samples that are unsuitable for meaningful artifact analysis.

\input{tables/supplementary/dataset_training_specs}

\subsection{Detailed Statistics of ViF-CoT-4K}
\label{subsec:detailed_statistics}
We further reveal the statistic details of ViF-CoT-4K and ViF-Bench, including a detailed report of the proportion of different types of artifacts annotated in ViF-CoT-4K (Table.~\ref{tab:annotation_distribution}), and technical details of the video generation model in ViF-CoT-4K and ViF-Bench (Table.~\ref{tab:gen_models}).

\subsection{Generated Video Examples}
\label{subsec:aigc_examples}
\begin{figure*}[h]
    \centering
    \includegraphics[width=0.92\textwidth]{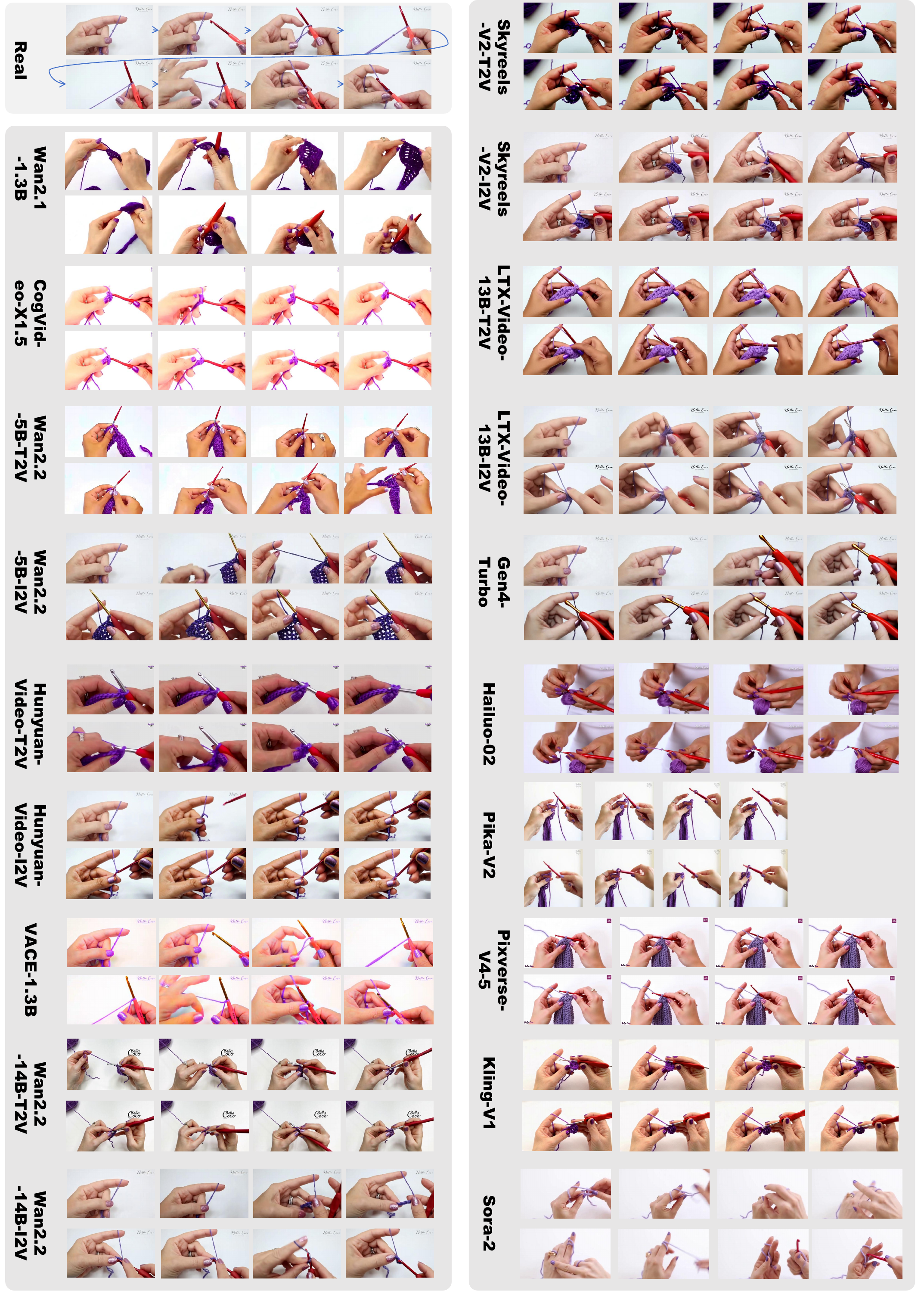}
    \vspace{-4.8mm}
    \caption{
        ViF-Bench Video Sample Examples-I
    }
    \vspace{-5mm}
    \label{fig:aigc_examples_I}
\end{figure*}

We demonstrate the quality of our dataset and benchmark by showing several examples randomly selected from the ViF-Bench (Figures~\ref{fig:aigc_examples_I}\&~\ref{fig:aigc_examples_II}). As shown in the image, fake samples in our dataset are generated by latest video generation models, and are closely aligned with their real counterparts to mitigate their gap in semantics and format.

\section{Analysis of Generalization Ablations.}
\label{sec:generalization_ablation}
\input{tables/generalization_ablation}
\noindent\textbf{Generalization Analysis: RL is the primary driver of cross-domain generalization.} We further isolate each component's contribution to generalization by evaluating under in-generator, cross-generator, and cross-dataset settings (Table~\ref{tab:generalization_ablation}). While SFT significantly improves in-domain accuracy, it also introduces mild overfitting, as cross-dataset accuracy drops from 68.34\% to 63.98\%. The subsequent RL stage recovers and extends generalization, yielding a +7.8\% cross-dataset accuracy gain over SFT while preserving in-domain performance. This suggests that RL encourages the model to discover intrinsic artifact cues rather than generator-specific patterns.

\section{Analysis of Artifact Cues}
\label{sec:artifact_cue_analysis}
To provide insight into what visual and temporal cues Skyra relies on for detection, we analyze the distribution of artifacts detected by Skyra on ViF-Bench, as shown in Table~\ref{tab:artifact_distribution}.

Among all detected artifacts, 82.8\% belong to \textit{Violation of Laws} while only 17.2\% are \textit{Low-Level Forgery}, indicating that Skyra primarily performs semantic reasoning about physical and logical inconsistencies rather than relying on low-level generator fingerprints. Within the \textit{Violation of Laws} category, \textit{Object Inconsistency} accounts for 28.1\%, with \textit{Shape Distortion} (15.2\%) being the single most dominant artifact type. This reflects a shared difficulty across current video generators in maintaining rigid-body constraints during dynamic scenes. \textit{Interaction Inconsistency} (10.0\%) and the remaining categories (44.7\%) further demonstrate that Skyra attends to diverse physical violations spanning object permanence, causality, and commonsense.

The relatively low proportion of \textit{Low-Level Forgery} (17.2\%) suggests that Skyra does not overfit to superficial texture or color cues, which tend to be generator-specific and less generalizable. Instead, its reliance on semantic-level violations aligns with how humans identify AI-generated content, contributing to its stronger cross-generator and cross-dataset generalization observed in Table~\ref{tab:generalization_ablation}.

\input{tables/supplementary/artifact_distribution}

\section{Analysis of Detection Capabilities}
\label{sec:analysis_of_baselines}
\subsection{Binary Detectors}
\label{subsec:binary_detectors}
\begin{figure}[t]
    \centering
    \includegraphics[width=0.45\textwidth]
    {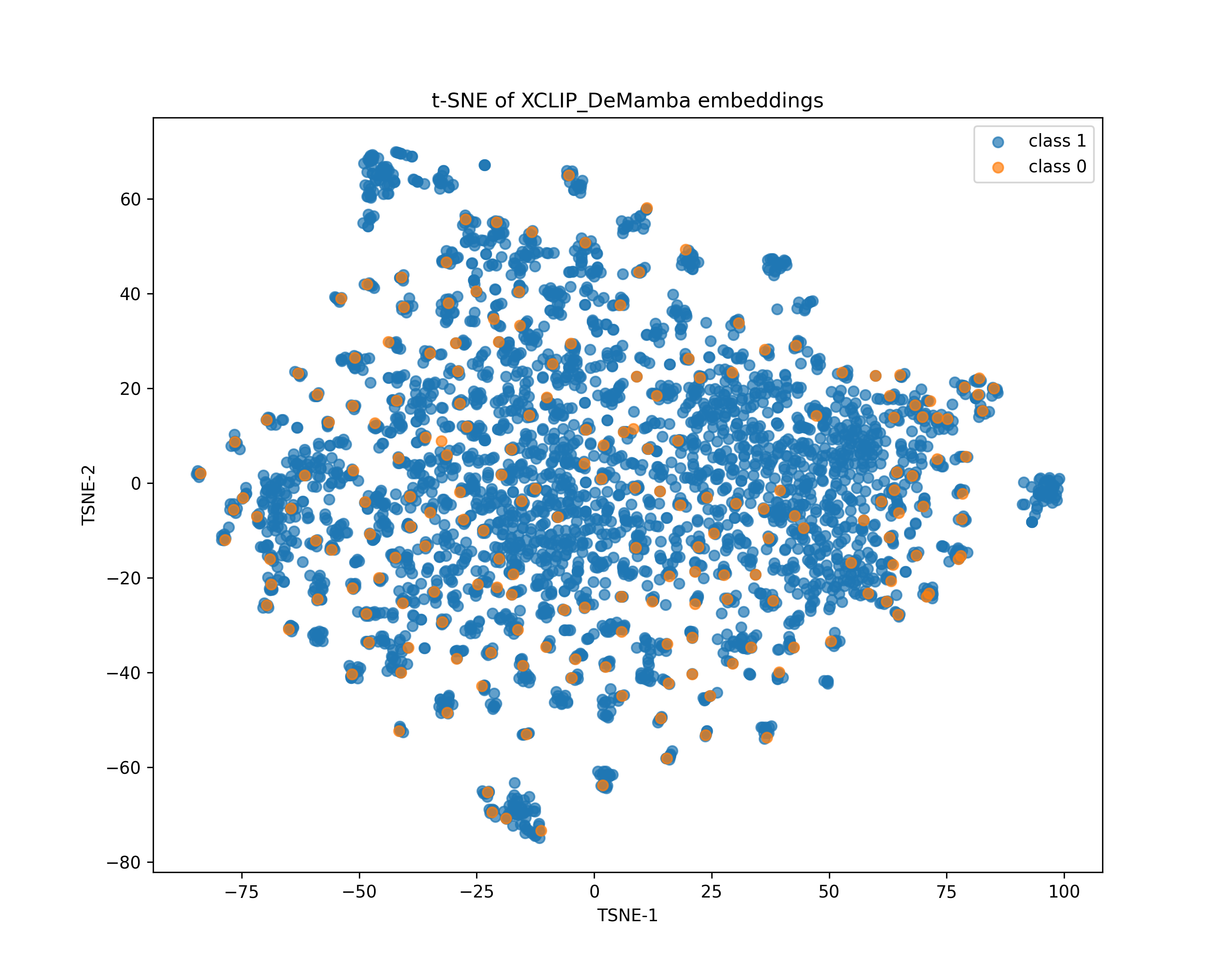}
    \vspace{-3.5mm}
    \caption{
        The T-SNE result of Demamba.
    }
    \vspace{-5mm}
    \label{fig:t_sne_Demamba}
\end{figure}

\begin{figure}[t]
    \centering
    \includegraphics[width=0.45\textwidth]
    {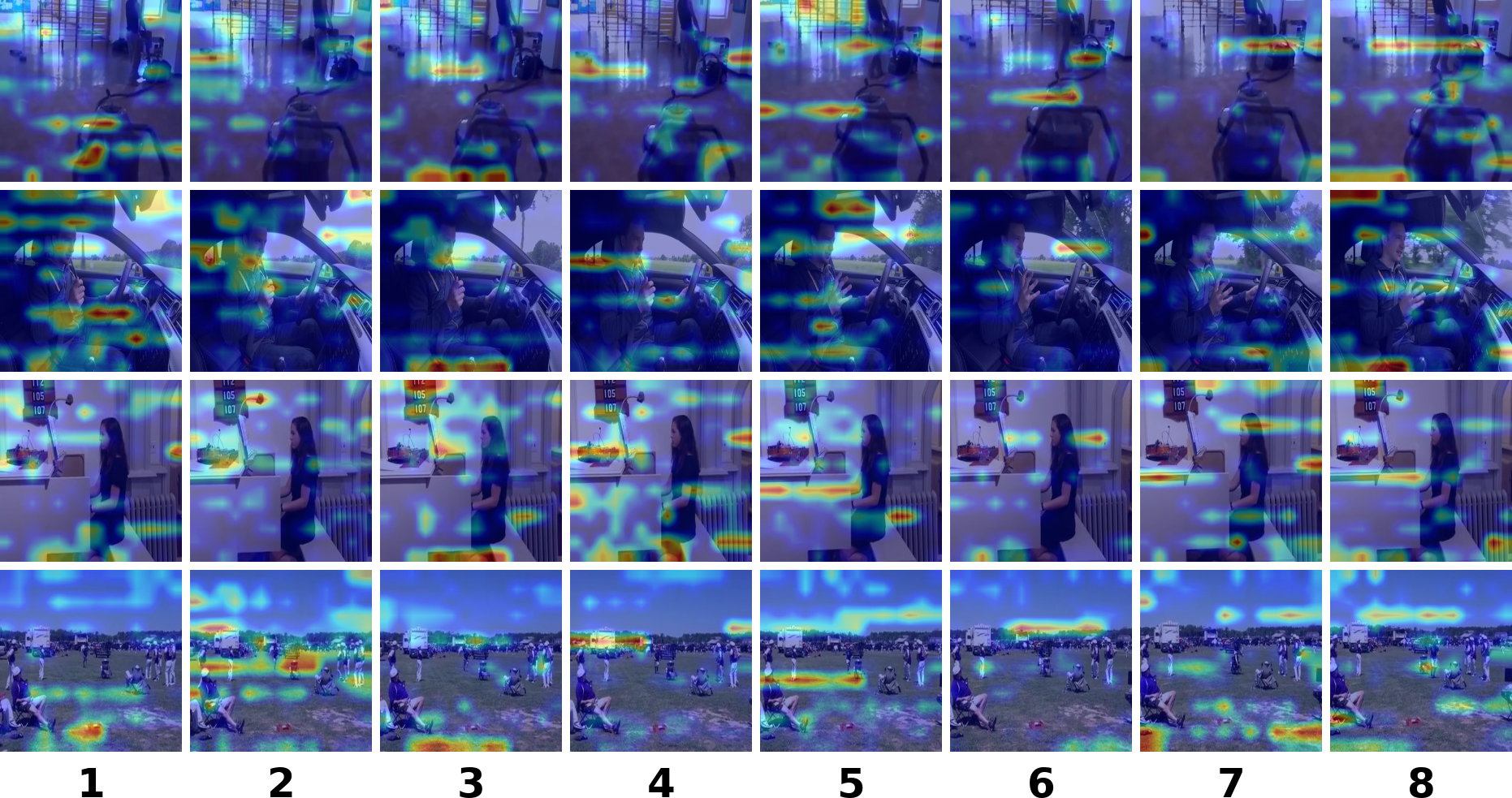}
    \vspace{-3.5mm}
    \caption{
        Visualization of Class Activation Maps (CAMs) produced by DeMamba on real video samples.
    }
    \vspace{-5mm}
    \label{fig:cam_Demamba}
\end{figure}

We take Demamba~\cite{chen2024demamba} and NSG-VD~\cite{zhang2025physics} as examples of underperforming classifiers that exhibit a strong tendency to label samples as fake.

\textbf{Demamba}: Through T-SNE visualization (Figure~\ref{fig:t_sne_Demamba}) and CAM heatmap analysis (Figure~\ref{fig:cam_Demamba}), we observe that in the T-SNE embedding space, fake and real samples are highly overlapping and difficult to separate. CAM heatmaps further reveal that, for real-labeled videos, the model consistently focuses on similar spatial locations across different samples, particularly in the third and fourth frames of the sequences. This suggests that
the model may be overly sensitive to fixed visual patterns or preferred spatial locations in the scene, rather than learning generalized content-based cues such as human motion or manipulation traces. Overall, the model does not attend to regions that are discriminative for authenticity, but rather to textures outside the main content of the frame, indicating limited generalization.

\begin{figure*}[!t]
    \centering
    \includegraphics[width=1\textwidth]{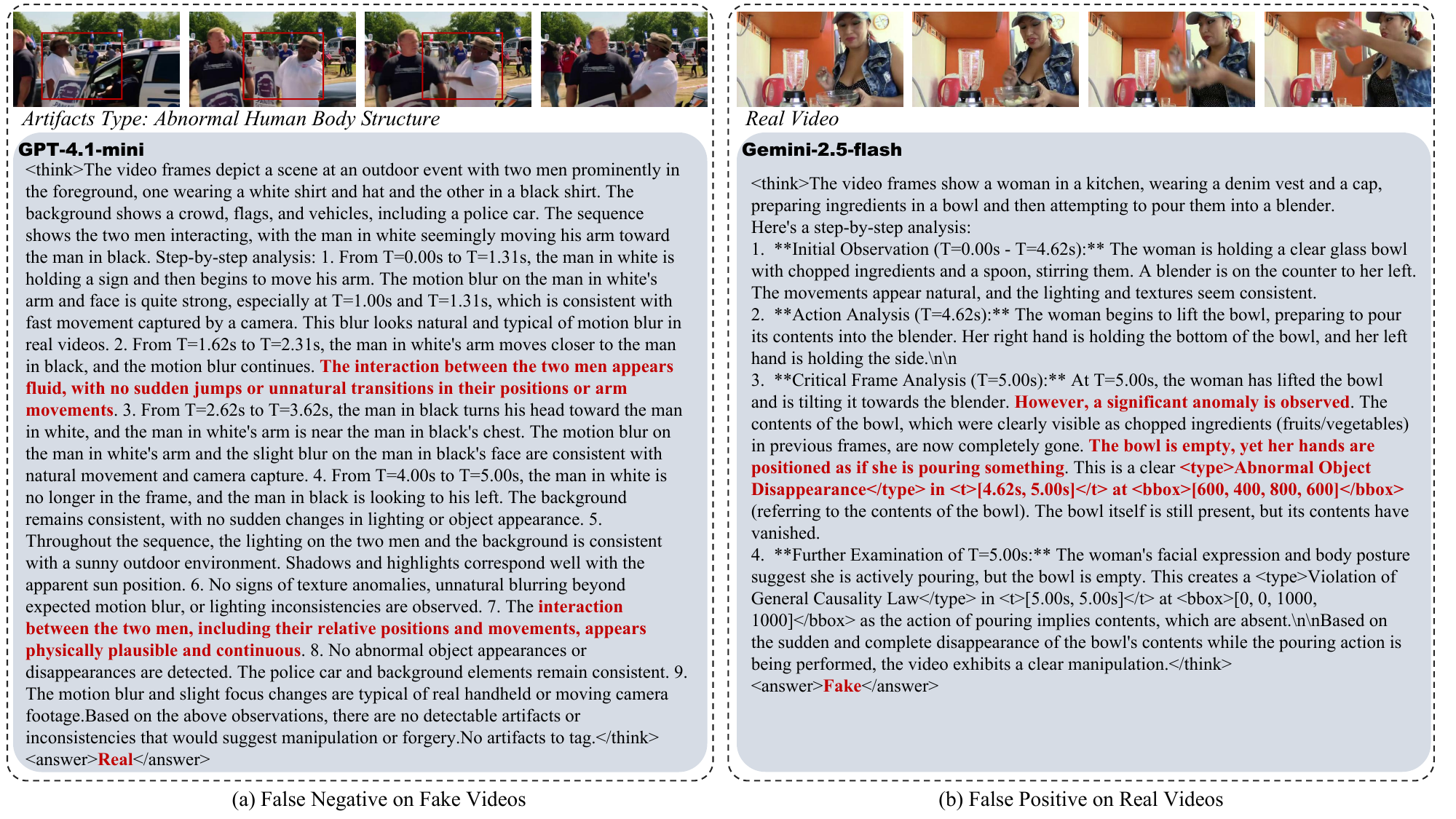}
    \vspace{-7mm}
    \caption{
        Response examples of off-the-shelf MLLMs.
    }
    \vspace{-5mm}
    \label{fig:examples_off_the_shelf_mllm}
\end{figure*}

\textbf{NSG-VD}: This classifier leverages a reference dataset, providing NSG feature baselines from real videos during inference. Specifically, the normalized spatiotemporal gradient (NSG) features of the reference data are used to model the distribution of real video dynamics, and the maximum mean discrepancy (MMD) between test videos and this reference distribution is computed. A test video is classified as AI-generated if its MMD exceeds a predefined threshold.

In our reproduction, the model achieved near-perfect AUROC on the validation set but behaved poorly on the test set, indicating a strong tendency to overfit. This overfitting may stem from the intrinsic sensitivity of NSG-based methods to subtle distributional shifts: the model struggles to generalize when the spatiotemporal dynamics of real videos deviate even slightly from those observed during training. When evaluating NSG-VD on the OOD GenVideo Benchmark, we consider it inappropriate to use the real samples within the GenVideo Benchmark as the reference dataset. Doing so would contradict the purpose of forgery detection and could introduce data leakage. Therefore, we retain the same reference dataset used during training. Similarly, in our robustness study, we do not use real samples under various degraded formats as reference data, because assuming access to degradation-specific real videos for each testing condition is impractical in real-world applications. This decision to reuse the training-time reference dataset may partly explain the suboptimal performance of NSG-VD in both experiments.

Overall, while the reference-dataset mechanism in NSG-VD provides strong detection capability when the distribution of the target data is known or partially accessible, its reliance on such reference information limits its effectiveness in OOD or noisy scenarios, where the real samples to be evaluated are either unavailable or should not be incorporated into the reference set in the first place.

\begin{figure*}[!t]
    \centering
    \includegraphics[width=1\textwidth]{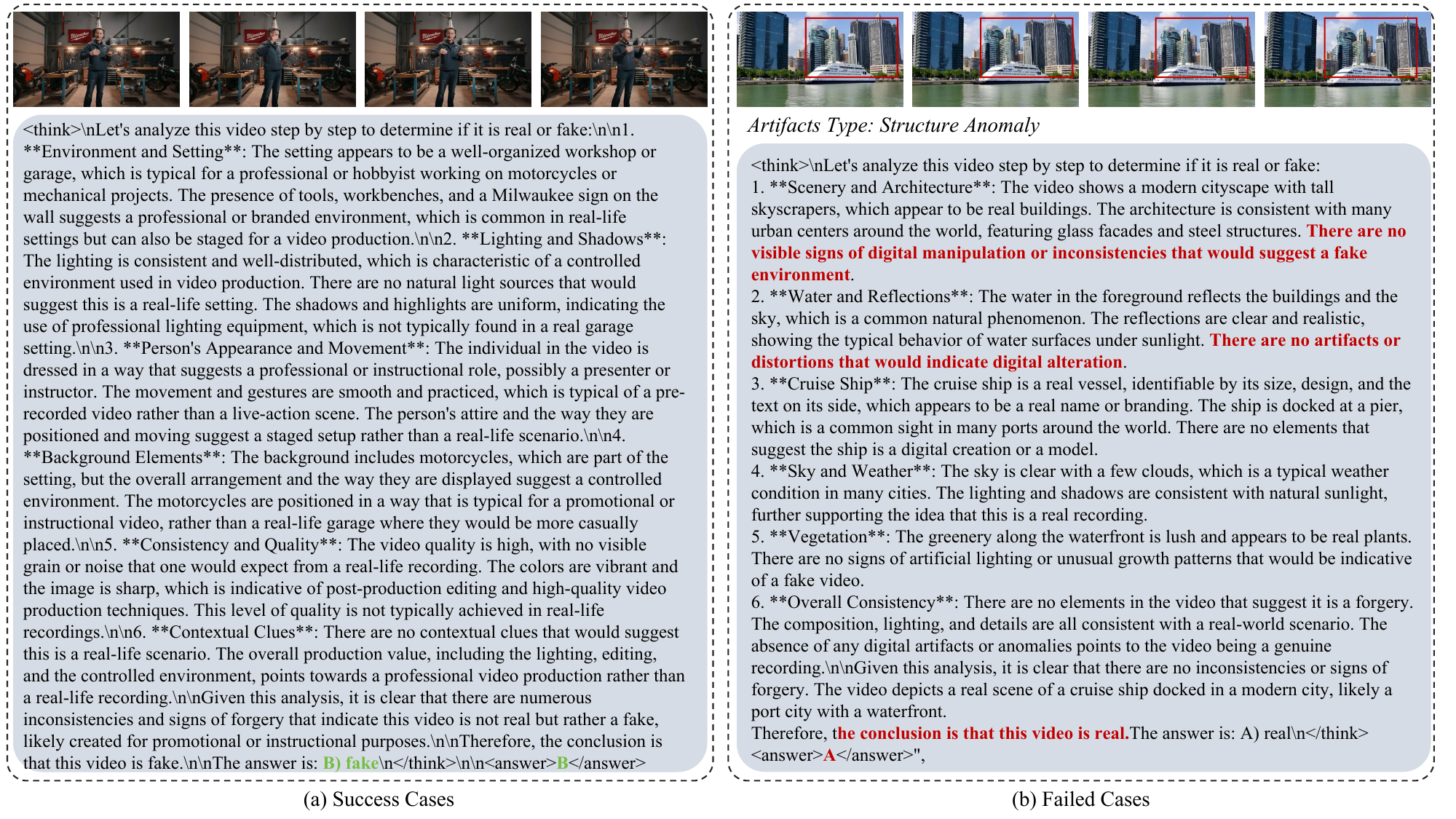}
    \vspace{-4mm}
    \caption{
        Response examples of existing MLLM-based detector, BusterX++~\cite{wen2025busterx++}.
    }
    \vspace{-5mm}
    \label{fig:examples_busterx}
\end{figure*}

\subsection{Off-the-Shelf MLLMs}
\label{subsec:off-the-self-MLLM}
Figure~\ref{fig:examples_off_the_shelf_mllm} provides concrete examples of the failure modes discussed in the main paper when directly prompting off-the-shelf MLLMs for AI-generated video detection. In panel (a), we show \emph{false negatives} on fake videos. Even with explicit chain-of-thought instructions, most models focus on high-level semantics and overall visual appeal (e.g., ``the scene looks natural'' or ``the movements are smooth'') while overlooking intrinsic forgery cues such as inconsistent geometry or physics-violating motion. As a result, they confidently classify clearly synthetic videos as real and provide rationales that largely describe the content instead of analyzing subtle spatiotemporal artifacts.

Panel (b) shows the opposite pattern. Models such as Gemini-2.5-flash~\cite{google_gemini2.5} tend to over-interpret natural video degradations, including compression artifacts, motion blur, and low-light noise, as evidence of forgery. In these cases, the model produces detailed yet incorrect explanations that attribute the degradations to ``AI generation'' rather than common acquisition or post-processing effects. This confirms our quantitative findings that off-the-shelf MLLMs tend to conflate quality with authenticity: they are sensitive to superficial visual cues but struggle to distinguish genuine forgery artifacts from benign imperfections in real-world videos.

\subsection{Existing MLLM-based Detectors}
\label{subsec:existing_mllm_detectors}
We further analyze BusterX++~\cite{wen2025busterx,wen2025busterx++}, a recent MLLM-based detector that adapts pretrained models for AIGC video detection. Figure~\ref{fig:examples_busterx} (a) shows a \emph{success case} where BusterX++ correctly identifies an AI-generated video. In such scenarios, the synthetic content exhibits obvious stylistic or aesthetic discrepancies from typical real videos (e.g., overly smooth textures or globally inconsistent lighting), which align well with the model’s training biases and allow it to reach the correct decision.

However, panels (b) highlight the limitations of relying primarily on global scene appearance.
Here, the AI-generated clip contains subtle but critical physics-violating artifacts, which humans readily notice. BusterX++, however, focuses on the overall coherence and visual quality of the scene and fails to attend to these localized spatiotemporal inconsistencies, leading to an incorrect ``real'' prediction. Together, these examples corroborate our main observation that current MLLM-based detectors behave more like general content describers: they emphasize superficial, distribution-level cues and natural degradations, but are not yet equipped to systematically discover and reason about intrinsic forgery artifacts that are crucial for reliable AI-generated video detection.

\section{Additional Examples}
\label{sec:additional_examples}
\subsection{Design of Prompts}
\label{subsec:design_of_prompt}
\begin{figure}[t]
    \centering
    \includegraphics[width=0.45\textwidth]{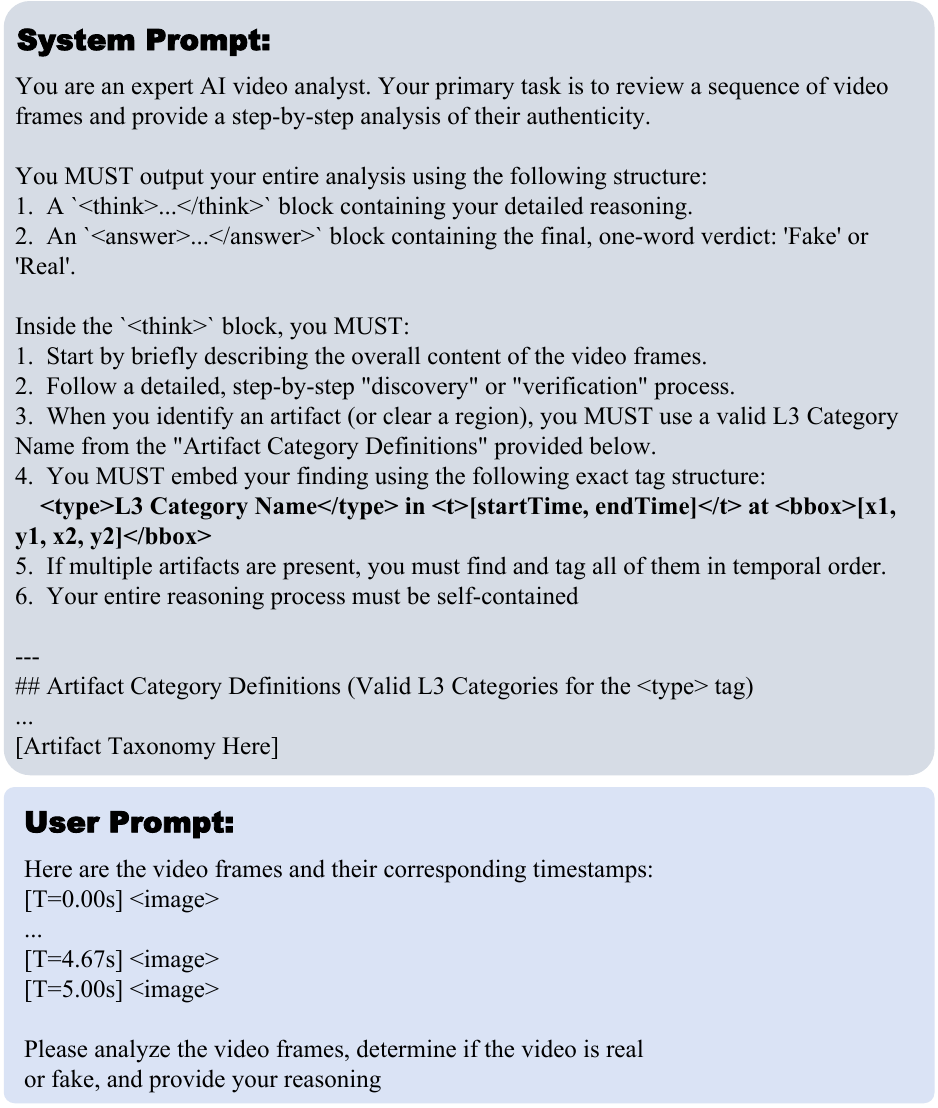}
    \vspace{-3.5mm}
    \caption{
        System prompt and user prompt design.
    }
    \vspace{-5mm}
    \label{fig:prompt_design}
\end{figure}

We specify the system and user prompt that Skyra uses in Figure~\ref{fig:prompt_design}. The system prompt specifies the model's role as an AI video analyst, clearly defines the output format (a \texttt{<think>} reasoning block followed by a one-word \texttt{<answer>} verdict), and constrains the reasoning to our artifact taxonomy, requiring that all findings be tagged with explicit categories, time spans, and bounding boxes. In contrast, the user prompt focuses on supplying multimodal evidence: we interleave sampled frames with their timestamps (e.g., ``[T=0.00s] \textless image\textgreater'' … ``[T=5.00s] \textless image\textgreater''), so that the model can reason over the evolution of the scene, align artifacts with precise temporal positions, and improve its ability to detect subtle, time-dependent inconsistencies.

\subsection{Examples of the Responses of Skyra}
\label{subsec:examples_of_skyra}
\begin{figure*}[h]
    \centering
    \includegraphics[width=0.92\textwidth]{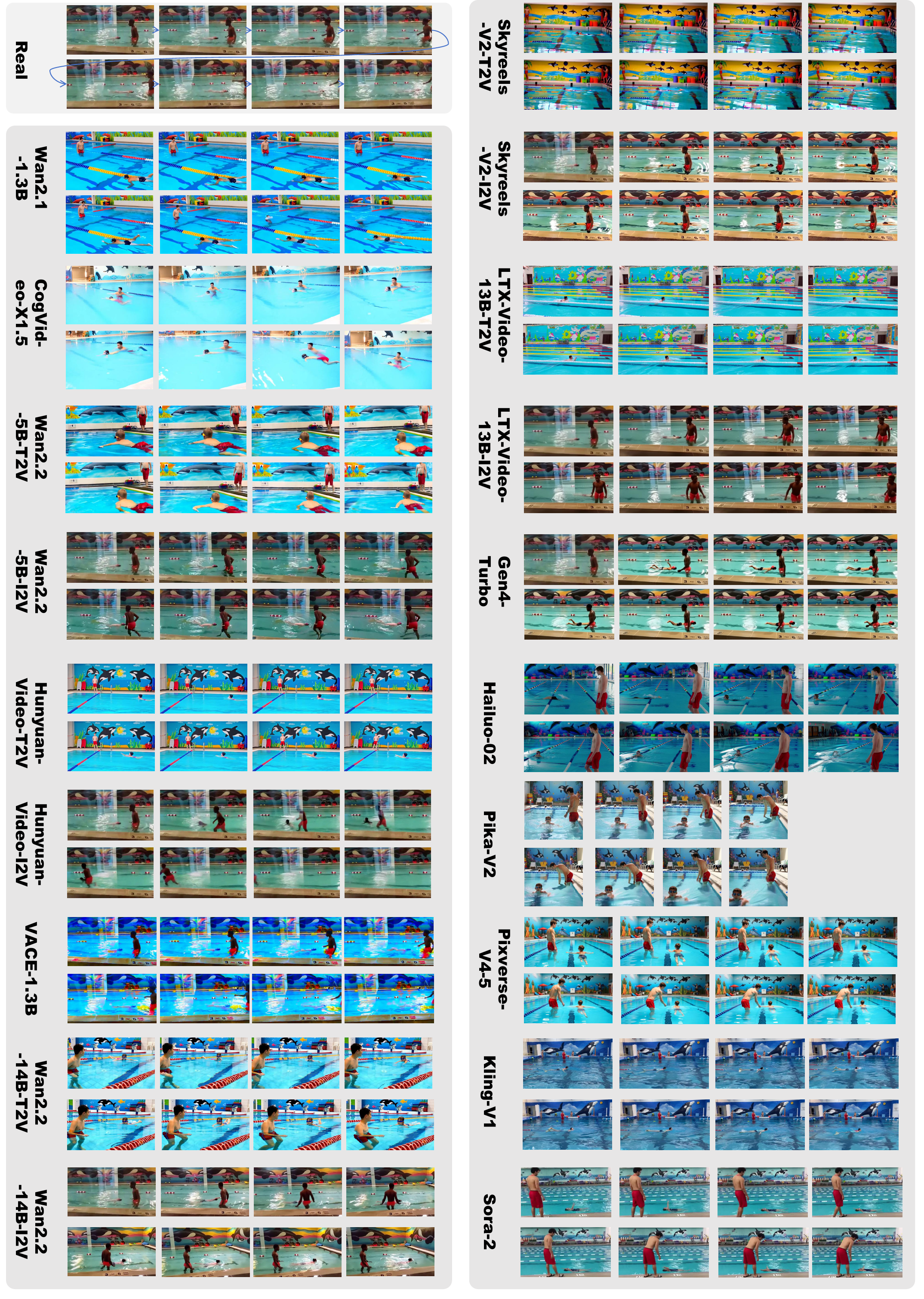}
    \vspace{-4.8mm}
    \caption{
        ViF-Bench Video Sample Examples-II
    }
    \vspace{-5mm}
    \label{fig:aigc_examples_II}
\end{figure*}

\begin{figure*}[h]
    \centering
    \includegraphics[width=1\textwidth]{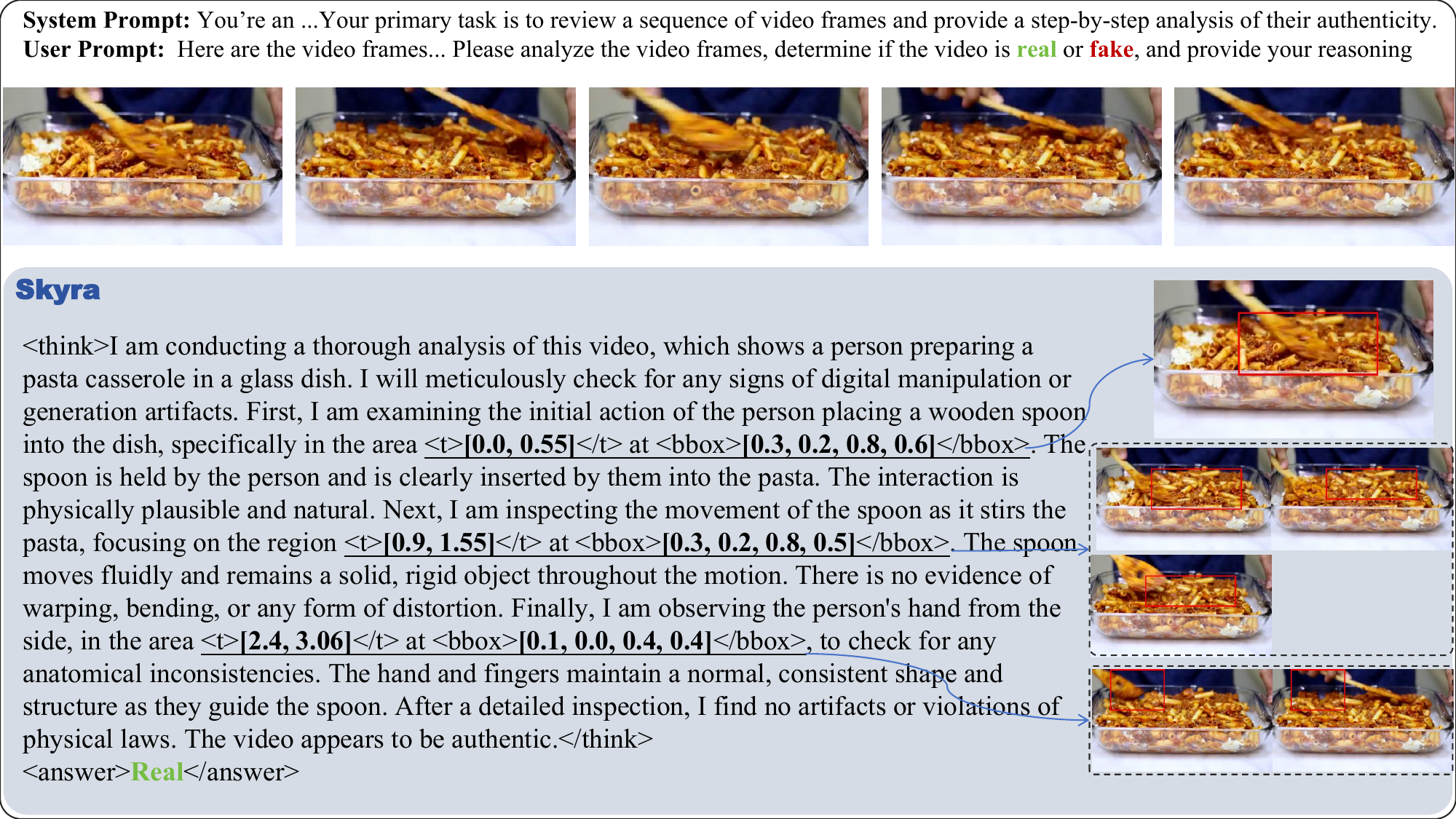}
    \vspace{-3mm}
    \caption{
        Skyra's Response Example on Real Videos, I
    }
    \vspace{-5mm}
    \label{fig:skyra_examples-I}
\end{figure*}

\begin{figure*}[h]
    \centering
    \includegraphics[width=1\textwidth]{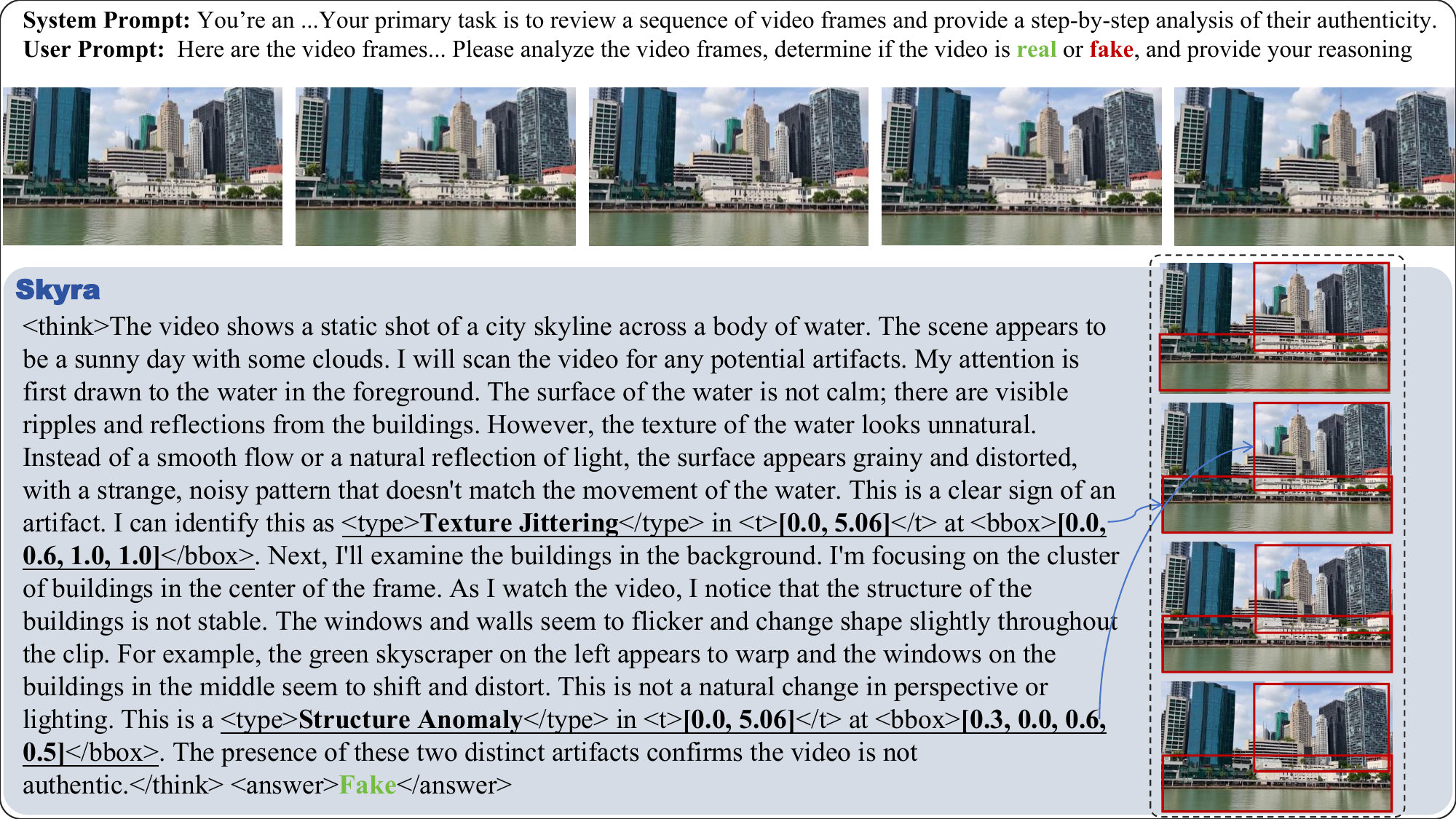}
    \vspace{-3mm}
    \caption{
        Skyra's Response Example on Fake Videos, Texture Anomaly-Structure Anomaly
    }
    \vspace{-5mm}
    \label{fig:skyra_examples-texture-anomaly}
\end{figure*}

\begin{figure*}[h]
    \centering
    \includegraphics[width=1\textwidth]{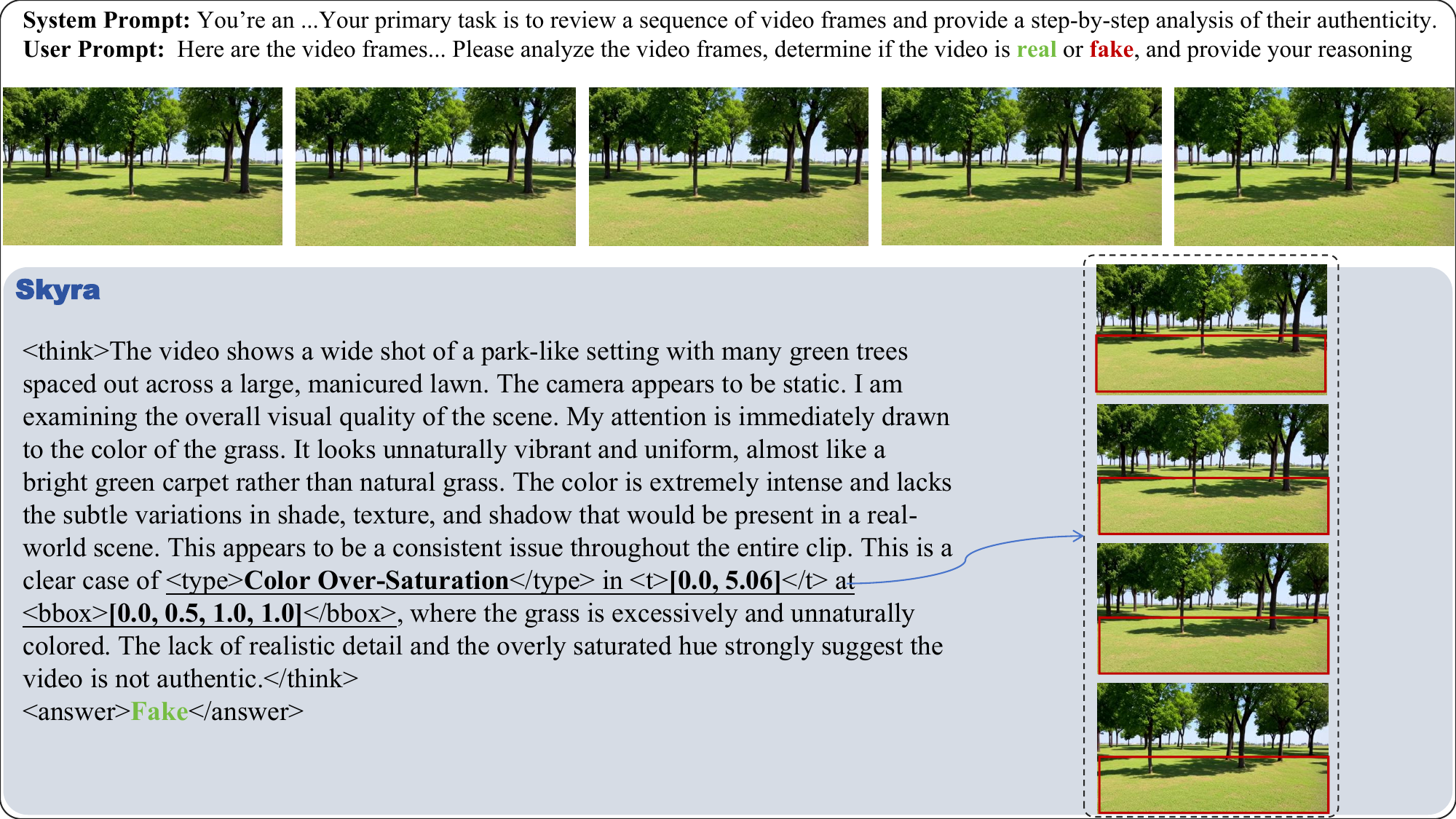}
    \vspace{-3mm}
    \caption{
        Skyra's Response Example on Fake Videos, Color \& Lighting Anomaly-Color Over-Saturation
    }
    \vspace{-5mm}
    \label{fig:skyra_examples-color-lighting-anomaly}
\end{figure*}

\begin{figure*}[h]
    \centering
    \includegraphics[width=1\textwidth]{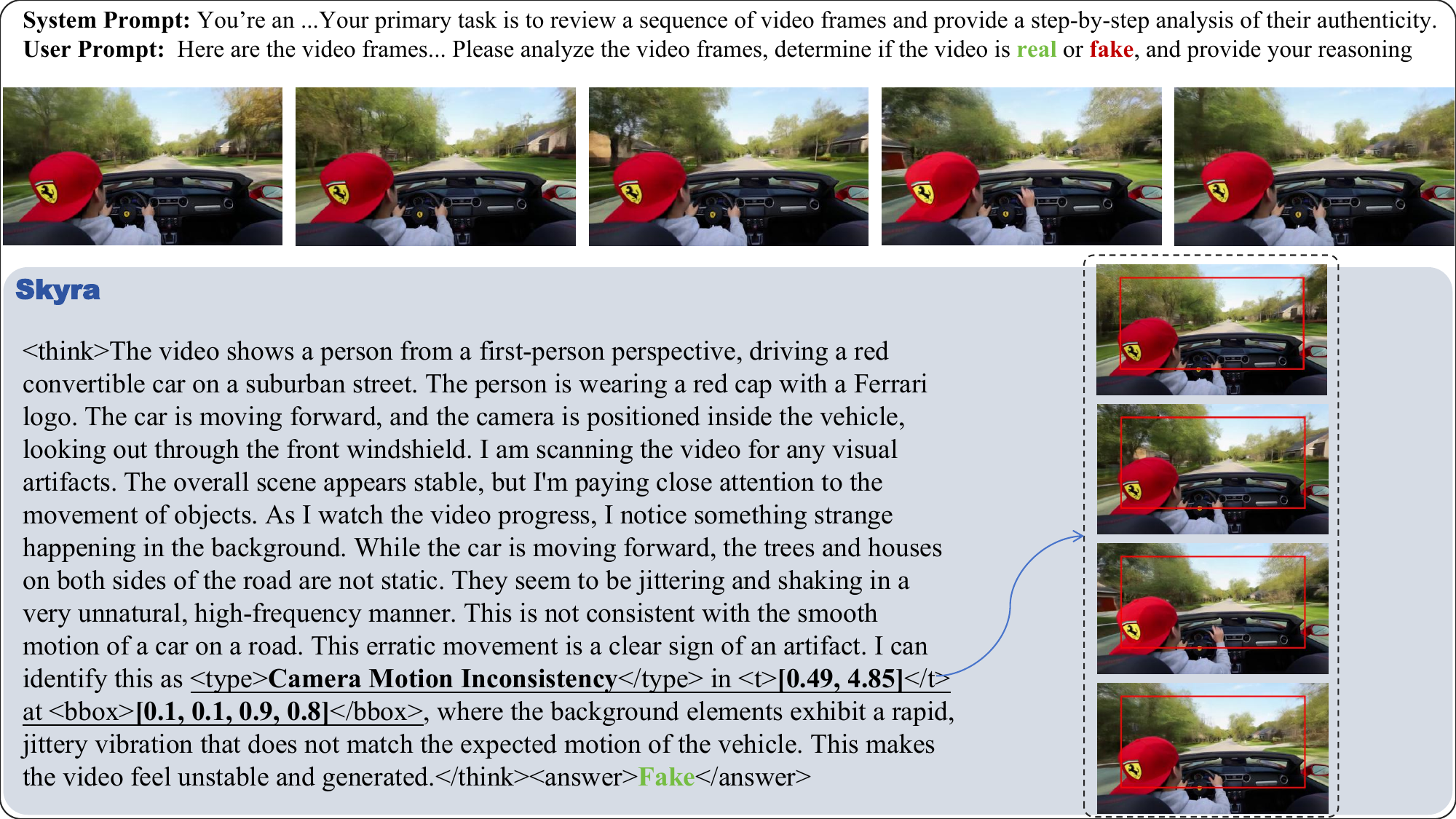}
    \vspace{-3mm}
    \caption{
        Skyra's Response Example on Fake Videos, Move Forgery-Camera Motion Inconsistency
    }
    \vspace{-5mm}
    \label{fig:skyra_examples-move-forgery}
\end{figure*}

\begin{figure*}[h]
    \centering
    \includegraphics[width=1\textwidth]{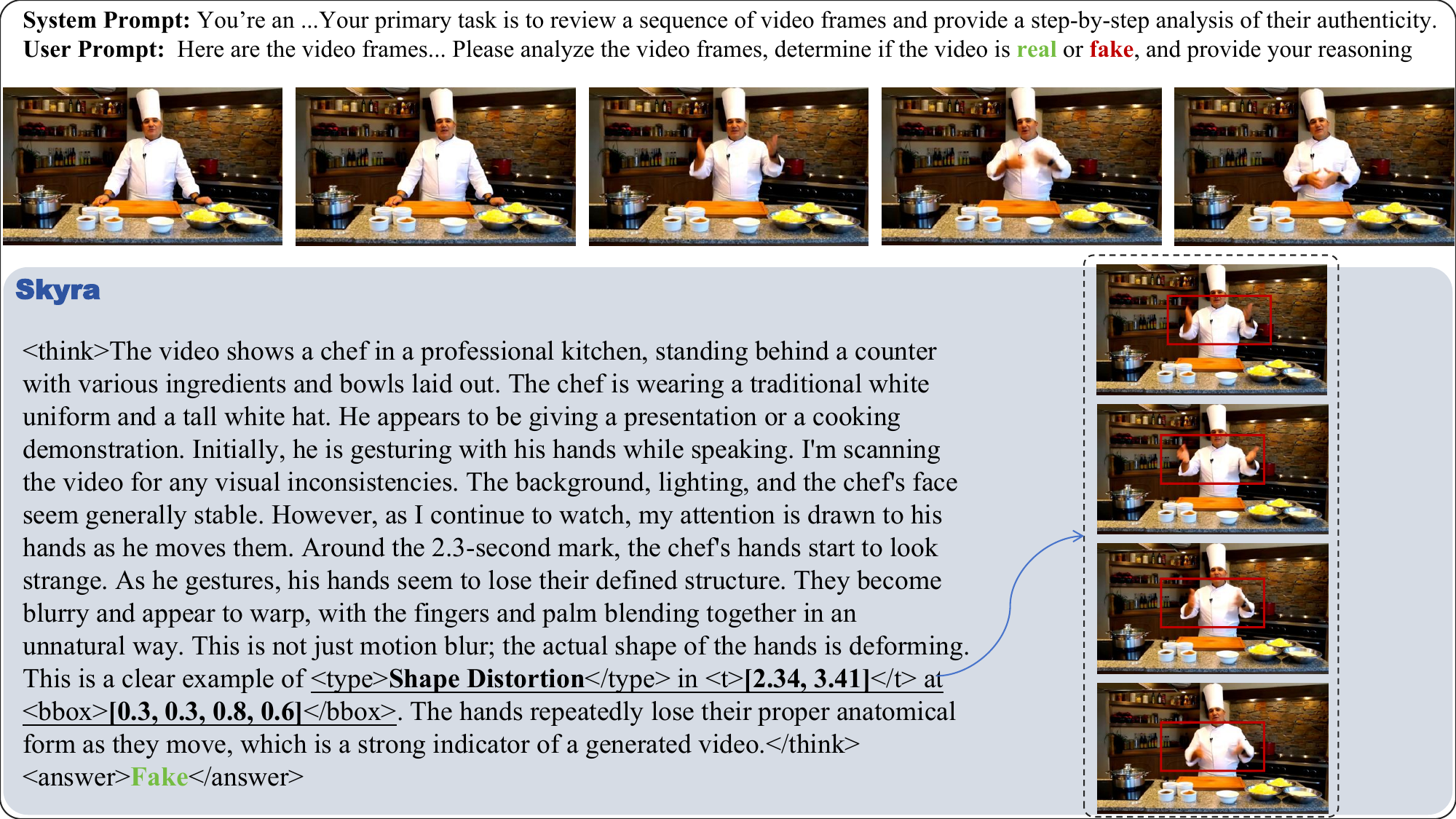}
    \vspace{-3mm}
    \caption{
        Skyra's Response Example on Fake Videos, Object Inconsistency-Shape Distortion
    }
    \vspace{-5mm}
    \label{fig:skyra_examples-object-inconsistency}
\end{figure*}

\begin{figure*}[h]
    \centering
    \includegraphics[width=1\textwidth]{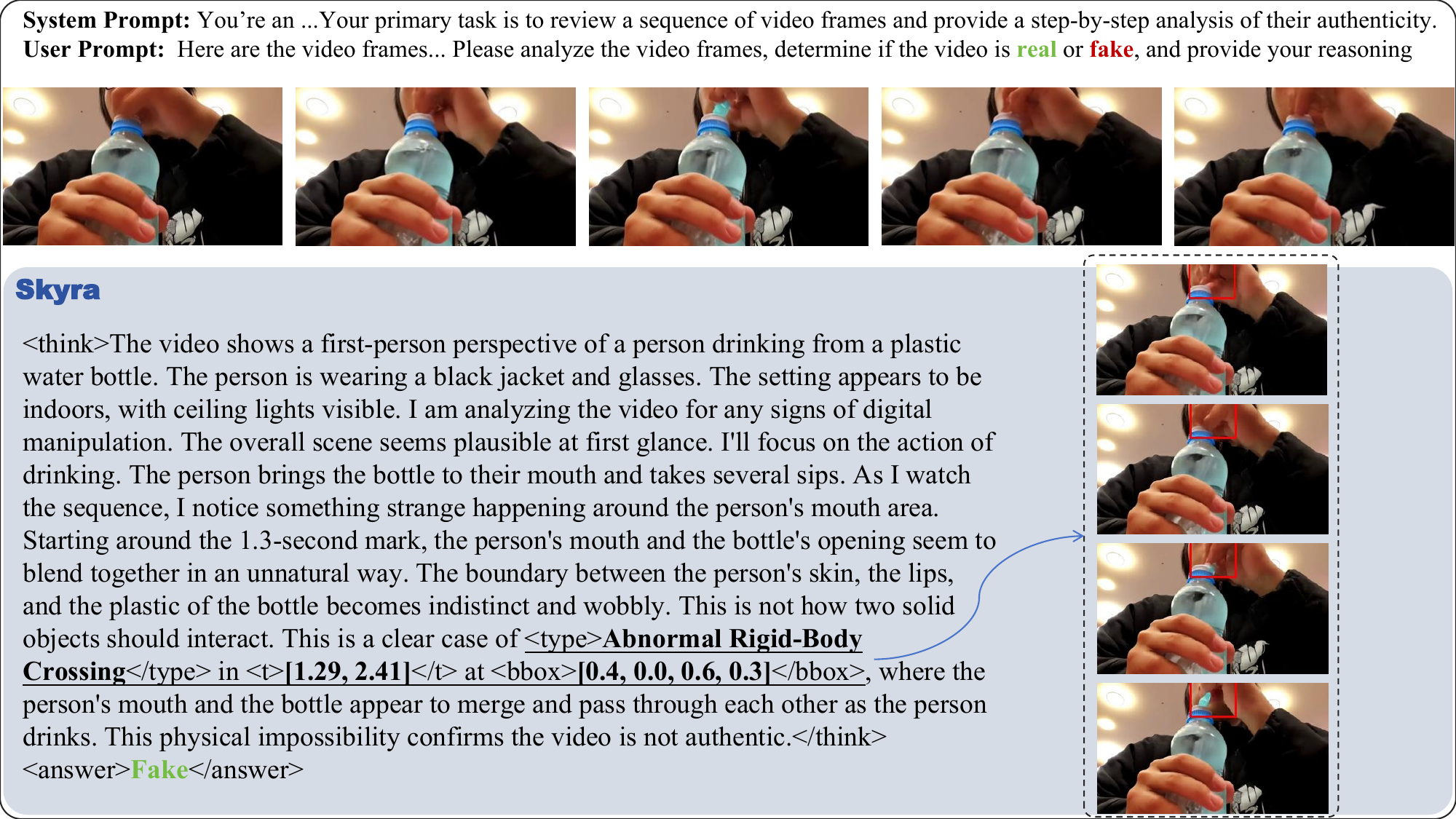}
    \vspace{-3mm}
    \caption{
        Skyra's Response Example on Fake Videos, Interaction Inconsistency-Abnormal Rigid-Body Crossing
    }
    \vspace{-5mm}
    \label{fig:skyra_examples-interaction-inconsistency}
\end{figure*}

\begin{figure*}[h]
    \centering
    \includegraphics[width=1\textwidth]{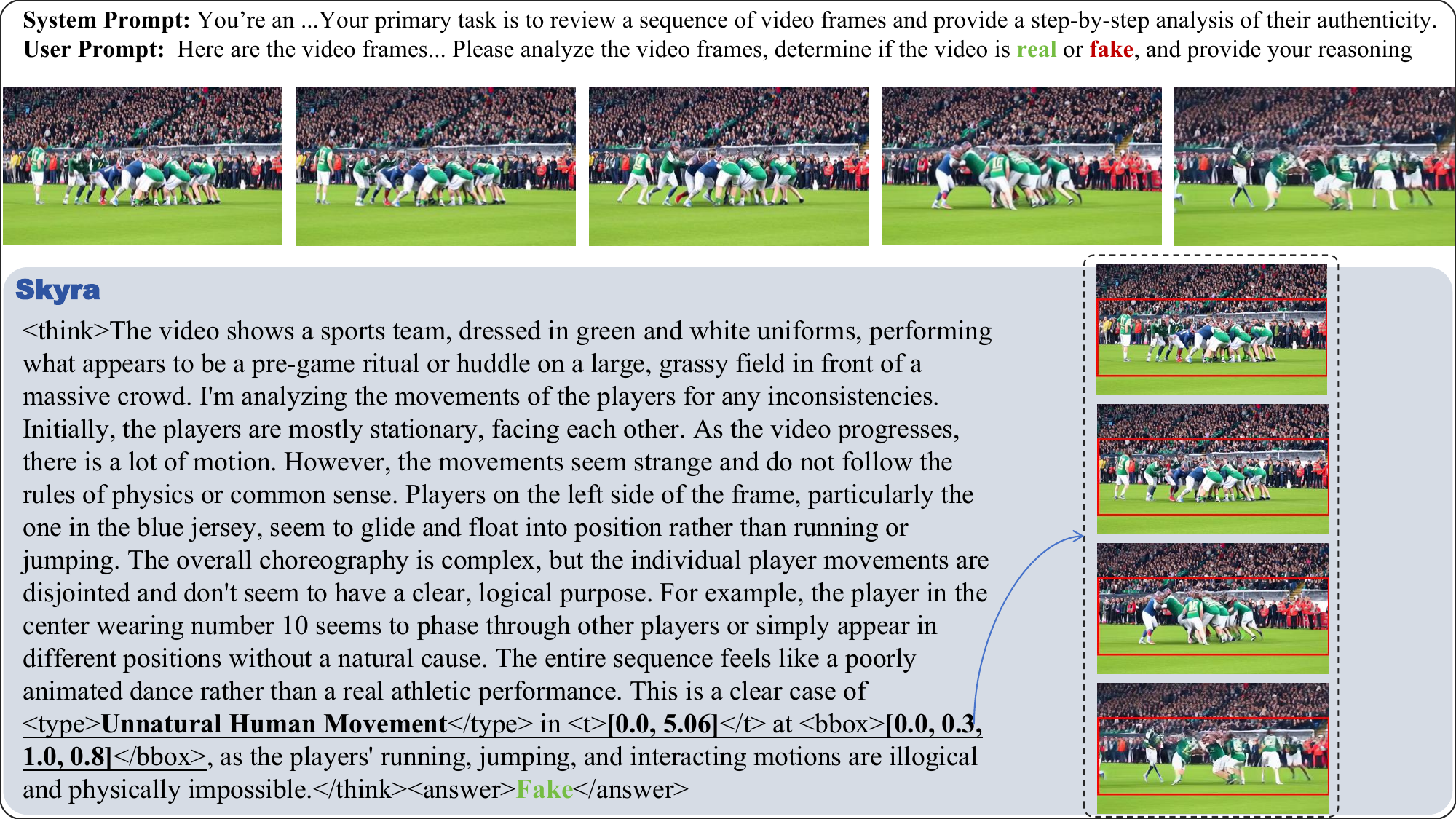}
    \vspace{-3mm}
    \caption{
        Skyra's Response Example on Fake Videos, Unnatural Movement-Unnatural Human Movement
    }
    \vspace{-5mm}
    \label{fig:skyra_examples-unnatural-movement}
\end{figure*}

\begin{figure*}[h]
    \centering
    \includegraphics[width=1\textwidth]{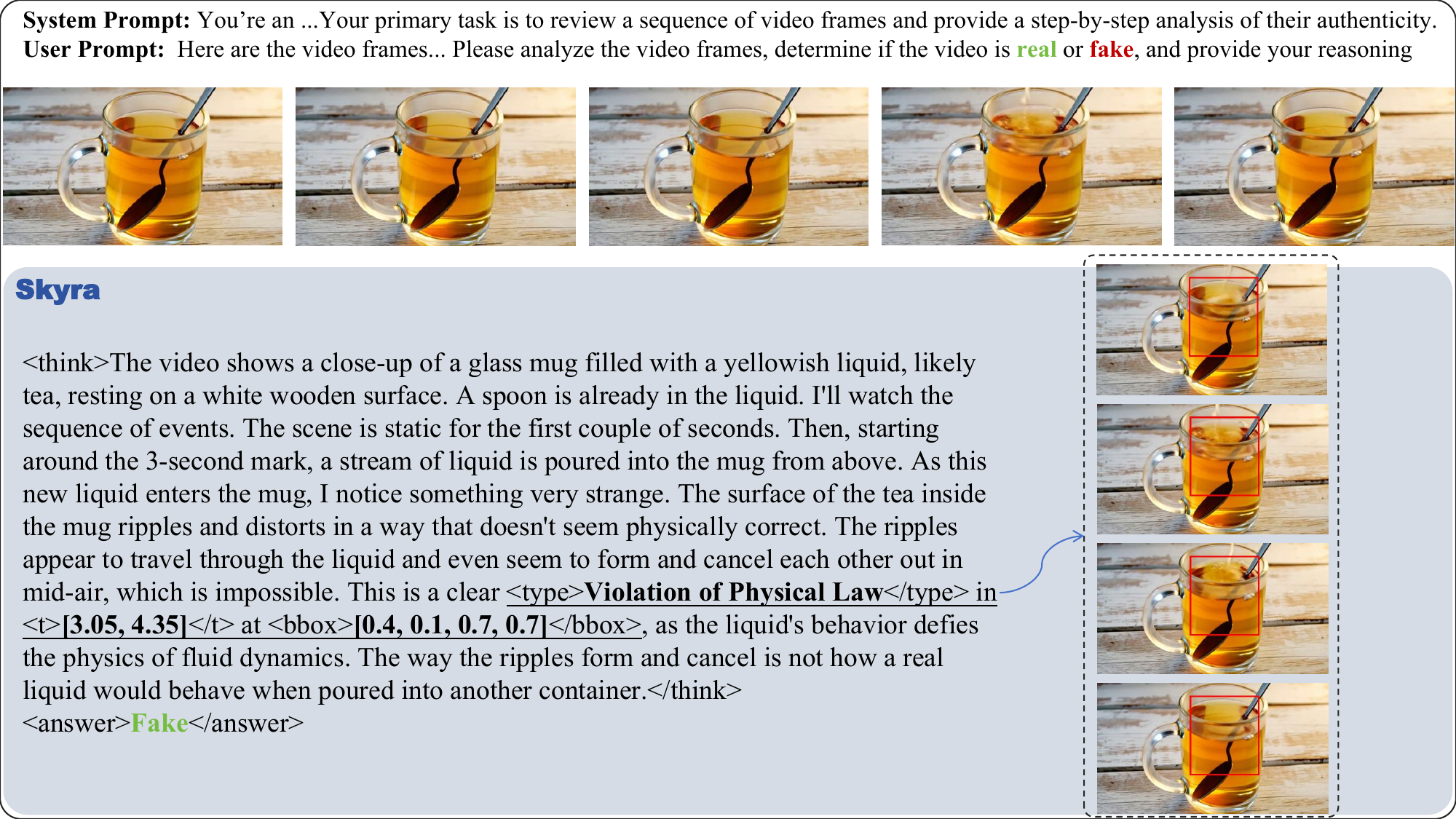}
    \vspace{-3mm}
    \caption{
        Skyra's Response Example on Fake Videos, Violation of Causality Law-Violation of Physical Law
    }
    \vspace{-5mm}
    \label{fig:skyra_examples-violation-of-causality-law}
\end{figure*}

\begin{figure*}[h]
    \centering
    \includegraphics[width=1\textwidth]{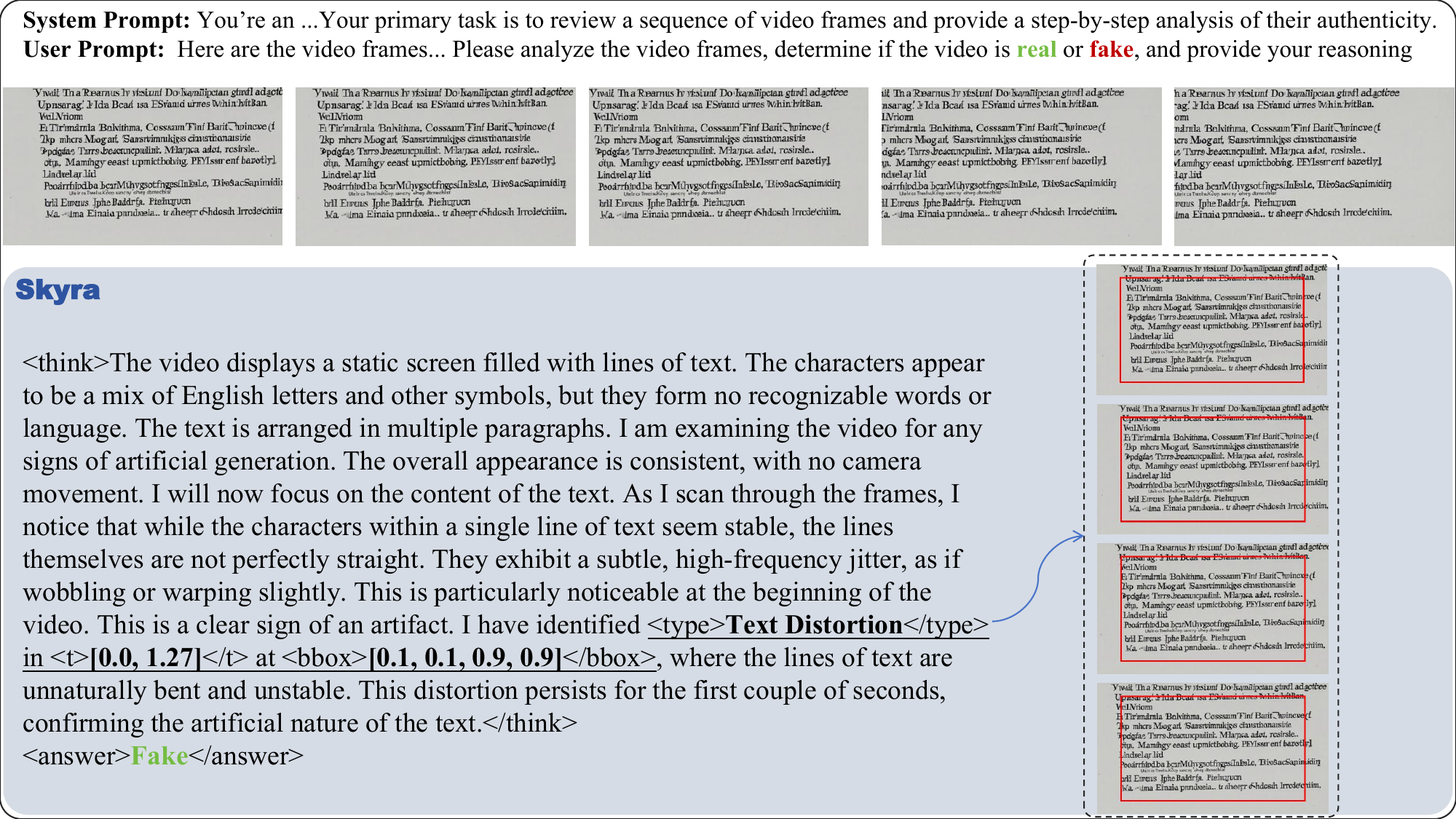}
    \vspace{-3mm}
    \caption{
        Skyra's Response Example on Fake Videos, Violation of Commonsense-Text Distortion
    }
    \vspace{-5mm}
    \label{fig:skyra_examples-violation-of-commonsense}
\end{figure*}

We provide the inference examples of Skyra on more samples in ViF-Bench. Figures~\ref{fig:skyra_examples-I} and \ref{fig:aigc_examples_II} demonstrate its responses when encountering real videos. Figure~\ref{fig:skyra_examples-texture-anomaly}-~\ref{fig:skyra_examples-violation-of-commonsense} exhibit different types of evidence that Skyra uses when determining that a video is AI-generated.

\section{Broader Impacts}
\label{subsec:broader_impacts}
Our work is motivated by the growing societal risks posed by AI-generated videos, including large-scale misinformation, impersonation, and erosion of trust in authentic media. By focusing on interpretable, artifact-centric detection, Skyra aims to provide not only predictions but also grounded visual evidence that can assist journalists, fact-checkers, regulators, and platform moderators in assessing the authenticity of suspicious content. The ViF-CoT-4K dataset and ViF-Bench further offer a standardized testbed for evaluating new detectors on diverse, up-to-date generators, which may contribute to more reliable and transparent AIGC safety tools.

At the same time, releasing a detailed artifact taxonomy, benchmark, and detector introduces dual-use concerns. In principle, insights into the failure modes of current detectors could inform future attempts to design more robust and evasive generative models. We believe that, on balance, the benefits of enabling the research community, civil society, and industry to build stronger and more interpretable defenses outweigh these risks. To mitigate potential misuse, our datasets contain only curated, non-sensitive content, and we emphasize that Skyra is intended to support human-in-the-loop verification rather than fully automated decision making or mass surveillance. We encourage downstream users to deploy our models and data in accordance with relevant regulations, to combine them with complementary safeguards such as provenance and watermarking, and to continuously stress-test detectors as the landscape of generative video models evolves.

\section{License}
\label{sec:license}
ViF-CoT-4K and ViF-Bench are provided to the community under CC BY 4.0 license. By downloading our dataset from our website or other sources, the user agrees to adhere to the terms of CC BY 4.0 and the licenses of the source datasets.  Licenses of the source datasets are listed in the Table~\ref{tab:license}.

\input{tables/supplementary/license}

\clearpage

\clearpage
\input{tables/supplementary/cot_annotation_portion}
\input{tables/supplementary/aigc_model_details}

\clearpage

\clearpage

\clearpage

\begin{figure*}[h]
    \centering
    \includegraphics[width=1\textwidth]{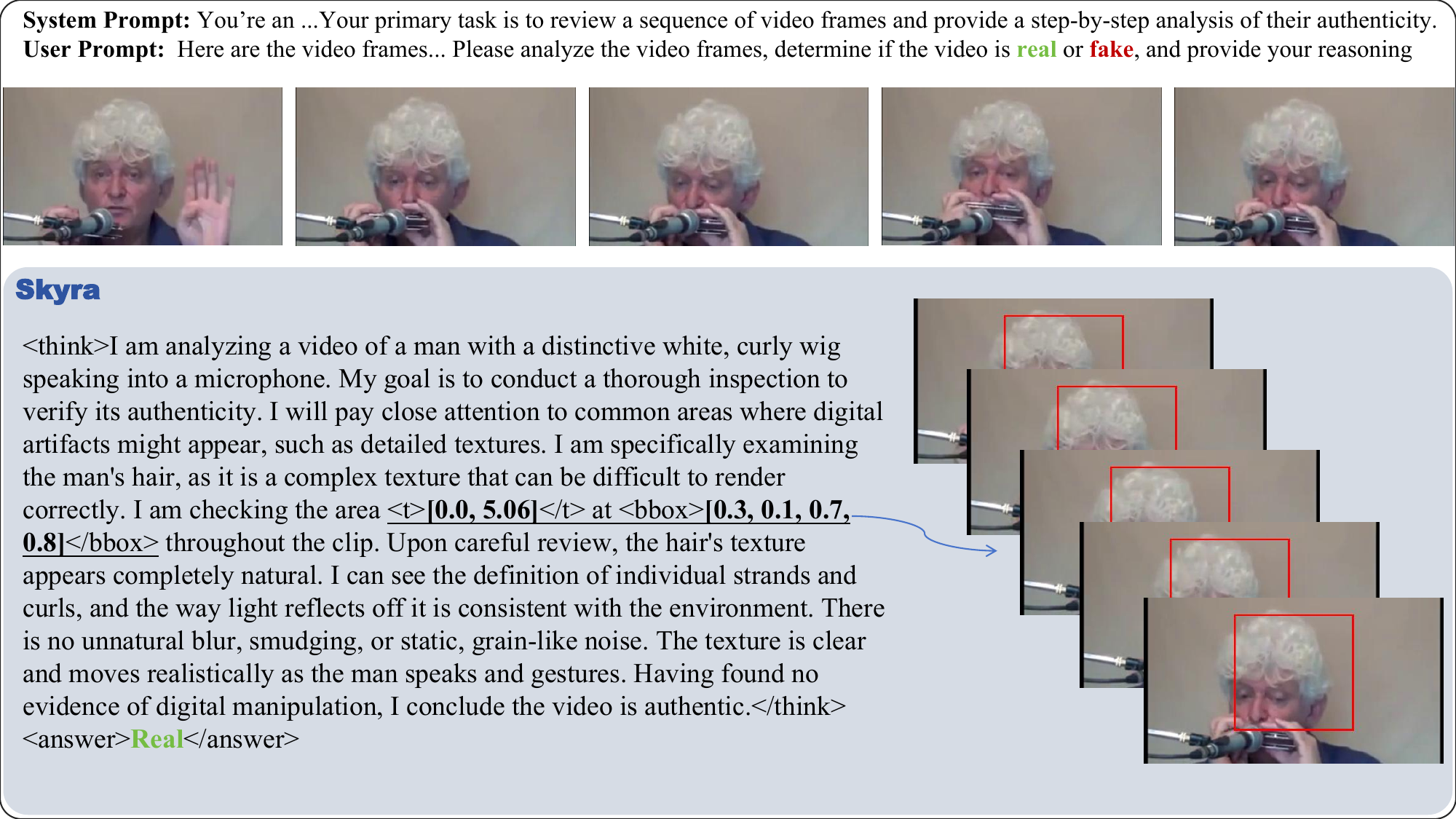}
    \vspace{-3mm}
    \caption{
        Skyra's Response Example on Real Videos, II
    }
    \vspace{-5mm}
    \label{fig:skyra_examples-II}
\end{figure*}

%% file: tables/supplementary/dataset_training_specs.tex
\begin{table}[h]
    \centering
    \small
    \setlength{\tabcolsep}{4pt}
    \renewcommand{\arraystretch}{1.1}
    \caption{Dataset and training specifications.}
    \vspace{-3mm}
    \resizebox{0.475\textwidth}{!}{
    \begin{tabular}{l|l}
        \toprule
        \multicolumn{2}{l}{\textit{Dataset Statistics}} \\
        \midrule
        \quad Video duration & 5 seconds (real \& fake, train \& test) \\
        \quad Total frames per video & 81 frames \\
        \quad Resolution & 256p (short side) \\
        \midrule
        \multicolumn{2}{l}{\textit{Training Settings}} \\
        \midrule
        \quad Sampled frames & 16 frames per video (uniformly sampled) \\
        \quad Input resolution & 256p (short side) \\
        \quad SFT learning rate & 1e-5 \\
        \quad SFT epochs & 5 \\
        \quad RL actor learning rate & 5e-7 \\
        \quad RL KL coefficient & 0.02 \\
        \quad Batch size per device & 1 \\
        \quad Hardware & 8$\times$ NVIDIA H200 GPUs \\
        \bottomrule
    \end{tabular}
    }
    \vspace{-3mm}
    \label{tab:dataset_training_specs}
\end{table}

%% file: tables/generalization_ablation.tex
\begin{table}[!t]
  \centering
  \small
  \setlength{\tabcolsep}{3pt}
  \caption{Generalization ablation study across in-generator, cross-generator, and cross-dataset settings on ViF-Bench and GenVideo.}
  \vspace{-3mm}
  \resizebox{0.475\textwidth}{!}{%
    \begin{tabular}{l|ccc|ccc|ccc}
      \toprule
      \multirow{2}{*}{Setting} & \multicolumn{3}{c|}{In-Generators} & \multicolumn{3}{c|}{Cross-Generators} & \multicolumn{3}{c}{Cross-Dataset} \\
      \cline{2-10}
       & Acc & R & F1 & Acc & R & F1 & Acc & R & F1 \\
      \midrule
      w/o CoT & 57.27 & 15.67 & 26.94 & 53.99 & 9.29 & 16.55 & 51.89 & 6.78 & 14.66 \\
      w/o SFT & 70.91 & 87.27 & 75.00 & 75.87 & \textbf{98.82} & 80.36 & \underline{68.34} & \underline{34.50} & \underline{46.00} \\
      Skyra-SFT & \underline{92.62} & \underline{89.49} & \underline{92.13} & \underline{88.95} & 82.42 & \underline{87.20} & 63.98 & 28.70 & 41.00 \\
      Skyra-RL & \textbf{93.58} & \textbf{93.22} & \textbf{93.45} & \textbf{89.84} & \underline{86.10} & \textbf{88.80} & \textbf{71.78} & \textbf{45.60} & \textbf{59.00} \\
      \bottomrule
    \end{tabular}%
  }
  \vspace{-5mm}
  \label{tab:generalization_ablation}
\end{table}

%% file: tables/supplementary/artifact_distribution.tex
\begin{table}[h]
  \centering
  \small
  \setlength{\tabcolsep}{2.5pt}
  \renewcommand{\arraystretch}{0.95}
  \caption{Distribution of artifacts detected by Skyra on ViF-Bench, organized by our hierarchical taxonomy.}
  \vspace{-3mm}
  \resizebox{0.475\textwidth}{!}{%
    \begin{tabular}{lllc}
      \toprule
      \textbf{L1 Category (Ratio)} & \textbf{L2 Category (Ratio)} & \textbf{L3 Category} & \textbf{Ratio} \\
      \midrule
      \multirow{4}{*}{\makecell[l]{Low-Level\\Forgery (17.2\%)}}
        & \multirow{3}{*}{\makecell[l]{Texture\\Anomaly (11.2\%)}}
          & Texture Jittering & 3.1\% \\
        & & Structure Anomaly & 3.5\% \\
        & & Others & 4.6\% \\
      \cline{2-4}
        & Others (6.0\%) & \dots & \dots \\
      \midrule
      \multirow{8}{*}{\makecell[l]{Violation of\\Laws (82.8\%)}}
        & \multirow{3}{*}{\makecell[l]{Object\\Inconsistency (28.1\%)}}
          & Shape Distortion & 15.2\% \\
        & & Abnormal Object Appearance & 5.8\% \\
        & & Others & 7.1\% \\
      \cline{2-4}
        & \multirow{3}{*}{\makecell[l]{Interaction\\Inconsistency (10.0\%)}}
          & Abnormal Multi-Object Merging & 2.7\% \\
        & & General Interaction Anomaly & 3.2\% \\
        & & Others & 4.1\% \\
      \cline{2-4}
        & Others (44.7\%) & \dots & \dots \\
      \bottomrule
    \end{tabular}%
  }
  \vspace{-3mm}
  \label{tab:artifact_distribution}
\end{table}

%% file: tables/supplementary/license.tex
\begin{table}[t]
    \centering
    \caption{License of source datasets in ViF-CoT-4K and ViF-Bench.}
    \vspace{-3mm}
    \resizebox{0.5\textwidth}{!}{
        \begin{tabular}{l|l}
            \toprule
            \textbf{Dataset} & \textbf{License}  \\
            \midrule
            Kinetics-400~\cite{kay2017kinetics} & CC BY 4.0  \\
            Panda-70M~\cite{chen2024panda} & Snap Inc. Non-Commercial Research \\
            HD-VILA-100M~\cite{xue2022advancing} & AGPL-3.0  \\
            \bottomrule
        \end{tabular}
    }
    \label{tab:license}
    \vspace{-5mm}
\end{table}

%% file: tables/supplementary/cot_annotation_portion.tex
\begin{table*}[t]
\centering
\caption{Hierarchical distribution of artifact categories (L1--L2--L3) in ViF-CoT-4K.}
\vspace{-2mm}
\resizebox{0.8\textwidth}{!}{
\begin{tabular}{l r | l r | l r}
\toprule
\textbf{L1 Category} & \textbf{Ratio} &
\textbf{L2 Category} & \textbf{Ratio} &
\textbf{L3 Category} & \textbf{Ratio} \\
\midrule

\multirow{6}{*}{\textbf{Low-Level Forgery}} 
& \multirow{6}{*}{17.2\%} 
& \multirow{3}{*}{Texture Anomaly} 
    & \multirow{3}{*}{11.2\%}
        & Structure Anomaly & 3.5\% \\
& & & & Texture Jittering & 3.1\% \\
& & & & Unnatural Blur & 3.6\% \\
\cline{3-6}

& & \multirow{2}{*}{Color \& Lighting Anomaly} 
    & \multirow{2}{*}{5.4\%}
        & Color Over-Saturation & 2.8\% \\
& & & & Lighting Inconsistency & 2.5\% \\
\cline{3-6}

& & \multirow{1}{*}{Move Forgery} 
    & \multirow{1}{*}{1.6\%}
        & Camera Motion Inconsistency & 1.6\% \\

\midrule

\multirow{24}{*}{\textbf{Violation of Laws}}
& \multirow{24}{*}{82.8\%}
& \multirow{5}{*}{Object Inconsistency}
    & \multirow{5}{*}{28.1\%}
        & Abnormal Object Disappearance & 3.6\% \\
& & & & Abnormal Object Appearance & 5.8\% \\
& & & & Person Identity Inconsistency & 1.1\% \\
& & & & General Object Identity Inconsistency & 2.4\% \\
& & & & Shape Distortion & 15.2\% \\
\cline{3-6}

& & \multirow{4}{*}{Interaction Inconsistency}
    & \multirow{4}{*}{10.0\%}
        & Abnormal Rigid-Body Crossing & 2.7\% \\
& & & & Abnormal Multi-Object Merging & 2.7\% \\
& & & & Abnormal Object Splitting & 1.4\% \\
& & & & General Interaction Anomaly & 3.2\% \\
\cline{3-6}

& & \multirow{3}{*}{Unnatural Movement}
    & \multirow{3}{*}{10.0\%}
        & Unnatural Human Movement & 6.6\% \\
& & & & Unnatural Animal Movement & 0.5\% \\
& & & & Unnatural General Object Movement & 2.9\% \\
\cline{3-6}

& & \multirow{2}{*}{Violation of Causality Law}
    & \multirow{2}{*}{6.90\%}
        & Violation of Physical Law & 4.1\% \\
& & & & Violation of General Causality Law & 2.8\% \\
\cline{3-6}

& & \multirow{3}{*}{Violation of Commonsense}
    & \multirow{3}{*}{27.8\%}
        & Abnormal Human Body Structure & 10.5\% \\
& & & & Abnormal General Object Structure & 3.2\% \\
& & & & Text Distortion & 14.1\% \\
\bottomrule
\end{tabular}
}
\label{tab:annotation_distribution}
\end{table*}

%% file: tables/supplementary/aigc_model_details.tex
\begin{table*}[t]
\centering
\small
\setlength{\tabcolsep}{3pt}
\caption{
Overview of video generation models used to synthesize forged samples in our dataset.
``Ref. Cond.'' denotes the typical conditioning modes (T2V: text-to-video, I2V: image-to-video, TI2V: text+image-to-video).
For commercial closed-source systems, parameter sizes are not publicly disclosed and thus marked as ``N/A (closed)''.
}
\resizebox{\textwidth}{!}{
\begin{tabular}{lcccccc}
\toprule
Model Name & Release Date & Parameter & Ref. Cond. & Inference & Sample Number & HyperLink\\
\midrule
Wan2.1-1.3B-T
  & 2025-02-25 & 1.3B
  & T2V & Local & 750 & \href{https://huggingface.co/Wan-AI/Wan2.1-T2V-1.3B}{Link} \\

CogVideoX1.5-T
  & 2024-08 & 5B
  & T2V & Local & 744 & \href{https://huggingface.co/zai-org/CogVideoX1.5-5BV}{Link}\\

CogVideoX1.5-I
  & 2024-08 & 5B
  & I2V & Local & 760 & \href{https://huggingface.co/zai-org/CogVideoX1.5-5B-I2V}{Link}\\

Wan2.2-TI2V-5B(T2V)
  & 2025-08-28 & 5B (MoE)
  & T2V & Local & 747 & \href{https://huggingface.co/Wan-AI/Wan2.2-TI2V-5B}{Link}\\

Wan2.2-TI2V-5B(I2V)
  & 2025-08-28 & 5B (MoE)
  & I2V & Local & 748 & \href{https://huggingface.co/Wan-AI/Wan2.2-TI2V-5B}{Link} \\

HunyuanVideo
  & 2024-12-03 & 13B
  & T2V & Local & 750 & \href{https://huggingface.co/tencent/HunyuanVideo}{Link} \\

HunyuanVideo-I2V
  & 2025-05-06 & 13B
  & I2V & Local & 968 & \href{https://huggingface.co/tencent/HunyuanVideo-I2V}{Link} \\

VACE-1.3B-T (Wan2.1-VACE-1.3B)
  & 2025-05-14 & 1.3B
  & T2V & Local & 165 & \href{https://huggingface.co/Wan-AI/Wan2.1-VACE-1.3B}{Link} \\

Wan2.2-T2V-A14B
  & 2025-08-28 & 14B (MoE)
  & T2V & Local & 165 & \href{https://huggingface.co/Wan-AI/Wan2.2-T2V-A14B}{Link} \\

Wan2.2-I2V-A14B
  & 2025-08-28 & 14B (MoE)
  & I2V & Local & 165 & \href{https://huggingface.co/Wan-AI/Wan2.2-I2V-A14B}{Link} \\

SkyReels-V2-T2V
  & 2025-04-21 & 14B
  & T2V & Local & 165 & \href{https://huggingface.co/Skywork/SkyReels-V2-T2V-14B-540P}{Link} \\

SkyReels-V2-I2V
  & 2025-04-21 & 14B
  & I2V & Local & 164 & \href{https://huggingface.co/Skywork/SkyReels-V2-I2V-14B-540P}{Link} \\

LTX-Video(T2V)
  & 2025-05-06 & 13B
  & T2V & Local & 165 & \href{https://huggingface.co/Lightricks/LTX-Video}{Link} \\

LTX-Video(I2V)
  & 2025-05-06 & 13B
  & I2V & Local & 165 & \href{https://huggingface.co/Lightricks/LTX-Video}{Link} \\

Gen4-Turbo
  & 2025-04 & N/A (closed)
  & I2V & API & 121 & \href{https://runwayml.com/research/introducing-runway-gen-4}{Link}\\

Hailuo-02
  & 2025-06-18 & N/A (closed)
  & T2V & API & 137 & \href{https://www.minimax.io/news/minimax-hailuo-02}{Link} \\

Pika-V2
  & 2025-08-15 & N/A (closed)
  & T2V & API & 151 & \href{https://pikalabs.org/pika-2-0/}{Link} \\

Pixverse-V4-5
  & 2025-05 & N/A (closed)
  & T2V & API & 152 & \href{https://app.pixverse.ai/onboard}{Link}\\

Kling-V1
  & 2024-06 & N/A (closed)
  & T2V & API & 141 & \href{https://klingai.com/}{Link} \\

Sora-2
  & 2024-02-15 & N/A (closed)
  & T2V & API & 150 & \href{https://openai.com/index/sora-2/}{Link} \\

\bottomrule
\end{tabular}
}

\vspace{2mm}
\label{tab:gen_models}
\raggedright
\footnotesize
\textbf{Note.} Dates are approximate and refer to the first public announcement or open release of the corresponding model family. For commercial systems with undisclosed architecture/size, ``Parameter'' is marked as ``N/A (closed)''.
\end{table*}